\documentclass[lettersize,journal]{IEEEtran}
\usepackage{amsmath,amsfonts}
\usepackage{algorithmic}
\usepackage{algorithm}
\usepackage{array}
\usepackage[caption=false,font=normalsize,labelfont=sf,textfont=sf]{subfig}
\usepackage{textcomp}
\usepackage{stfloats}
\usepackage{url}
\usepackage{verbatim}
\usepackage{graphicx}
\usepackage{cite}
\graphicspath{{img/}}
\usepackage{multirow}

\usepackage[utf8]{inputenc} 
\usepackage[T1]{fontenc}    
\usepackage{hyperref}       
\usepackage{booktabs}       
\usepackage{nicefrac}       
\usepackage{microtype}      
\usepackage{xcolor}         
\usepackage{colortbl}
\usepackage{amssymb}
\usepackage{mathtools}
\usepackage{amsthm}
\usepackage{mathrsfs}
\usepackage{bbm}
\usepackage{tasks}

\usepackage{wrapfig}
\usepackage[capitalize,noabbrev]{cleveref}
\usepackage{enumerate}
\usepackage{parskip}

\theoremstyle{plain}
\newtheorem{theorem}{Theorem}[section]
\newtheorem{proposition}[theorem]{Proposition}
\newtheorem{lemma}[theorem]{Lemma}
\newtheorem{corollary}[theorem]{Corollary}
\theoremstyle{definition}
\newtheorem{definition}[theorem]{Definition}

\theoremstyle{remark}
\newtheorem{remark}[theorem]{Remark}

\providecommand{\calM}{\mathcal{M}}
\providecommand{\calN}{\mathcal{N}}
\providecommand{\calA}{\mathcal{A}}

\providecommand{\sym}[1]{\mathcal{S}^{#1}}
\providecommand{\spd}[1]{\mathcal{S}^{#1}_{++}}
\providecommand{\cho}[1]{\mathcal{L}_{+}^{#1}}
\providecommand{\tril}[1]{\mathcal{L}^{#1}}
\providecommand{\bbR}[1]{\mathbb {R}^{#1}}
\providecommand{\bbRplus}{\mathbb {R}_{+}}
\providecommand{\bbRscalar}{\mathbb {R}}
\providecommand{\cinf}{C^{\infty}}
\providecommand{\orth}[1]{\mathrm{O}({#1})}

\providecommand{\rieexp}{\operatorname{Exp}}
\providecommand{\rielog}{\operatorname{Log}}
\providecommand{\diffphi}[1]{\phi_{*,#1}}
\providecommand{\diffphiinv}[1]{\phiinv_{*,#1}}
\providecommand{\diffmlog}[1]{\operatorname{mlog}_{*,#1}}
\providecommand{\diffmln}[1]{\operatorname{mln}_{*,#1}}
\providecommand{\diffmgexp}[1]{\phi_{ma *,#1}}
\providecommand{\pt}[2]{\Gamma_{#1 \rightarrow #2}}
\providecommand{\scrL}{\mathscr{L}}
\providecommand{\bbD}{\mathbb {D}}
\providecommand{\ln}{\operatorname{ln}}
\providecommand{\mln}{\operatorname{mln}}
\providecommand{\cln}{\phi_{cln}}
\providecommand{\clnchart}{\varphi_{ln}}

\providecommand{\mlog}{\operatorname{mlog}}

\providecommand{\mexp}{\phi_{\mathrm{mexp}}}
\providecommand{\mgexp}{\phi_{\mathrm{ma}}}

\providecommand{\mlnMul}{\odot_{mln}}
\providecommand{\clnMul}{\odot_{cln}}
\providecommand{\mlogMul}{\odot_{mlog}}

\providecommand{\mlnMulScalar}{\circledast_{mln}}

\providecommand{\mlogMulScalar}{\circledast_{mlog}}

\providecommand{\diag}{\operatorname{diag}}
\providecommand{\rieExp}[1]{\operatorname{Exp}_{#1}}

\providecommand{\phiinv}{\phi^{-1}}
\providecommand{\phiMulScalar}{\circledast_{\phi}}
\providecommand{\gphi}{g^{\phi}}
\providecommand{\phiMul}{\odot_{\phi}}
\providecommand{\diff}{\operatorname{d}}
\providecommand{\tr}{\operatorname{tr}}
\providecommand{\galem}{g^{\mathrm{ALE}}}
\providecommand{\glem}{g^{\mathrm{LE}}}
\providecommand{\glcm}{g^{\mathrm{LC}}}
\providecommand{\gcm}{g^{\mathrm{C}}}
\providecommand{\geuc}{g^{\mathrm{E}}}

\providecommand{\gphi}{g^{\phi}}
\providecommand{\dlem}{d^{\mathrm{LE}}}
\providecommand{\dlcm}{d^{\mathrm{LC}}}

\providecommand{\dalem}{d^{\mathrm{ALE}}}
\providecommand{\dphi}{d^{\phi}}

\providecommand{\fm}{{\mathrm{FM}}}
\providecommand{\det}{{\operatorname{det}}}
\providecommand{\gbiparamlem}{g^{(a,b)\text{-LE}}}
\providecommand{\gbiparamalem}{g^{(a,b)\text{-ALE}}}
\providecommand{\gbiparamaE}{g^{(a,b)\text{-E}}}
\providecommand{\glcm}{g^{\text{LC}}}

\providecommand{\balpha}{\boldsymbol{\alpha}}
\providecommand{\rmE}{\mathrm{E}}
\providecommand{\rmF}{\mathrm{F}}

\providecommand{\biparamLEM}{(a,b)\text{-LEM}}
\providecommand{\biparamALEM}{(a,b)\text{-ALEM}}
\providecommand{\biparam}{(a,b)}
\providecommand{\bfst}{\mathbf{ST}}

\providecommand{\ie}{\textit{i.e., }}
\providecommand{\vsacle}{\vspace{0pt}}
\providecommand{\etal}{\textit{et al.}}
\newcommand{\na}{\textcolor{gray}{N/A}}%

\providecommand{\spec}{\operatorname{spec}}

\providecommand{\gyr}{\operatorname{gyr}}
\providecommand{\gyrinner}[2]{\left\langle #1, #2 \right\rangle_{\mathrm{gr}}}
\providecommand{\gyrnorm}[1]{\left\| #1 \right\|_{\mathrm{gr}}}
\providecommand{\gyrdist}{{\mathrm{d}}_{\mathrm{gry}}}

\crefname{equation}{Eq.}{Eqs.}
\Crefname{equation}{Equation}{Equations}
\crefname{figure}{Fig.}{Figs.}
\Crefname{figure}{Figure}{Figures}
\crefname{table}{Tab.}{Tabs.}
\Crefname{table}{Table}{Tables}
\crefname{algocf}{Alg.}{Algs.}
\Crefname{algocf}{Algorithm}{Algorithms}
\crefname{section}{Sec.}{Secs.}
\Crefname{section}{Section}{Sections}
\crefname{appendix}{App.}{Apps.}
\Crefname{appendix}{Appendix}{Appendices}
\crefname{suplement}{Supp.}{Supps.}
\Crefname{suplement}{Suplement}{Suplements}

\crefname{theorem}{Thm.}{Thms.}
\Crefname{theorem}{Theorem}{Theorems}
\crefname{lemma}{Lem.}{Lems.}
\Crefname{lemma}{Lemma}{Lemmas}
\crefname{definition}{Def.}{Defs.}
\Crefname{definition}{Definition}{Definitions}
\crefname{corollary}{Cor.}{Cors.}
\Crefname{corollary}{Corollary}{Corollaries}
\crefname{remark}{Rem.}{Rems.}
\Crefname{remark}{Remark}{Remarks}
\crefname{proposition}{Prop.}{Props.}
\Crefname{proposition}{Proposition}{Propositions}
\crefname{proof}{Pr.}{Prs.}
\Crefname{proof}{Proof}{Proofs}

\crefname{enumi}{Case}{Cases}
\Crefname{enumi}{Case}{Cases}

\begin{document}

\title{Adaptive Log-Euclidean Metrics for SPD Matrix Learning}


\markboth{Journal of \LaTeX\ Class Files,~Vol.~14, No.~8, August~2021}%
{Shell \MakeLowercase{\textit{et al.}}: A Sample Article Using IEEEtran.cls for IEEE Journals}

\author{Ziheng~Chen,
        Yue~Song\textsuperscript{*},
        Tianyang~Xu,
        Zhiwu~Huang,
        Xiao-Jun~Wu,
        and
        Nicu~Sebe
        \thanks{This work was partly supported by the MUR PNRR project FAIR (PE00000013) funded by the NextGenerationEU, the EU Horizon project ELIAS (No. 101120237), and a donation from Cisco. The authors also gratefully acknowledge the financial support from the China Scholarship Council
        (CSC).}
        \thanks{Ziheng Chen, Yue Song, and Nicu Sebe are with the Department
        of Information Engineering and Computer Science, University of Trento,
        Trento, Italy.
        E-mail: ziheng\_ch@163.com, songyue19960927@gmail.com, niculae.sebe@unitn.it.
        (Corresponding author: Yue Song)}
        \thanks{Tianyang~Xu and Xiao-Jun~Wu are with the School of Artificial Intelligence and Computer Science, Jiangnan University, Wuxi, China.
        E-mail: {tianyang.xu; wu\_xiaojun}@jiangnan.edu.cn.}
        \thanks{Zhiwu~Huang is with the School of Electronics and Computer Science, University of Southampton, Southampton, U.K.
        E-mail: zhiwu.huang@soton.ac.uk.}
        \thanks{This paper has supplementary downloadable material available at \url{http://ieeexplore.ieee.org}, provided by the authors. The material includes preliminaries, implementation details, and proofs.}
        }


\maketitle

\begin{abstract}
    Symmetric Positive Definite (SPD) matrices have received wide attention in machine learning due to their intrinsic capacity to encode underlying structural correlation in data. 
    Many successful Riemannian metrics have been proposed to reflect the non-Euclidean geometry of SPD manifolds.
    However, most existing metric tensors are fixed, which might lead to sub-optimal performance for SPD matrix learning, especially for deep SPD neural networks.
    To remedy this limitation, we leverage the commonly encountered pullback techniques and propose Adaptive Log-Euclidean Metrics (ALEMs), which extend the widely used Log-Euclidean Metric (LEM).
    Compared with the previous Riemannian metrics, our metrics contain learnable parameters, which can better adapt to the complex dynamics of Riemannian neural networks with minor extra computations.
    We also present a complete theoretical analysis to support our ALEMs, including algebraic and Riemannian properties.
    The experimental and theoretical results demonstrate the merit of the proposed metrics in improving the performance of SPD neural networks.
    The efficacy of our metrics is further showcased on a set of recently developed Riemannian building blocks, including Riemannian batch normalization, Riemannian Residual blocks, and Riemannian classifiers.
\end{abstract}

\section{Introduction}
\label{sec:intro}

\IEEEPARstart{T}he Symmetric Positive Definite (SPD) matrices are ubiquitous in statistics, supporting a diversity of scientific areas, such as medical imaging~\cite{chakraborty2018statistical,das2018sparse,chakraborty2020manifoldnet}
, signal processing \cite{yair2019parallel,brooks2019riemannian,kobler2022spd,ju2024deep}, elasticity \cite{moakher2006averaging, guilleminot2012generalized}, question answering \cite{lopez2021vector,nguyen2022gyro}, graph and node classification \cite{zhao2023modeling}, and computer vision \cite{huang2017riemannian,li2017high,wang2018discriminant,qiao2019deep,nguyen2021geomnet,song2021approximate,nguyen2022gyrovector,song2022fast,wei2022discrete,chen2024liebn,chen2024rmlr}.
Despite the ability to capture data variations, SPD matrices cannot simply interact as points in the Euclidean space, which becomes the main challenge in practice.
To guarantee the manifoldness, several Riemannian metrics have been proposed, including Affine-Invariant Metric (AIM) \cite{pennec2006riemannian}, Log-Euclidean Metric (LEM) \cite{arsigny2005fast}, and Log-Cholesky Metric (LCM) \cite{lin2019riemannian}, to name a few.
Equipped with these metrics, many Euclidean methods could be generalized into the domain of the Riemannian manifold \cite{wang2012covariance, huang2015log, huang2015face, harandi2018dimensionality,chen2021hybrid}. It is essential to clarify that there are also some metric learning methods in SPD manifolds \cite{huang2015log,harandi2018dimensionality}.
However, the metrics these methods learned are distance functions induced by existing Riemannian metrics.
In contrast, this paper focuses on Riemannian metrics, which are more fundamental than the metric learning methods mentioned above.

Recently, inspired by the vivid progress of deep learning \cite{hochreiter1997long,krizhevsky2012imagenet,he2016deep}, several deep networks were developed on the SPD manifold \cite{huang2017riemannian,chakraborty2018statistical,brooks2019riemannian,chakraborty2020manifoldnet,pan2022matt,wang2022learning,wang2022dreamnet,nguyen2022gyrovector,nguyen2022gyro,chen2023riemannian,nguyen2023building,wang2024spd,chen2024liebn,ju2024deep,chen2024rmlr}.
Although different network structures are designed, the theoretical foundations of these methods are all built upon Riemannian metrics on the SPD manifold.
Therefore, the design of the Riemannian metric is significantly important for the efficacy of the learning algorithms.
However, most metric tensors in the existing popular Riemannian metrics on the SPD manifold are fixed, which could undermine the expressibility of the associated geometry. 
After analyzing several existing Riemannian metrics on SPD manifolds, we find that the pullback is a commonly used tool, which can be intuitively viewed as a bijection preserving Riemannian properties.
For instance, \cite{thanwerdas2022theoretically} explained AIM as the pullback metric from a left-invariant metric on the Cholesky manifold.
In \cite{thanwerdas2023n}, the authors generalized LEM by the pullback of the vanilla LEM.
In \cite{lin2019riemannian}, the authors proposed LCM by the pullback from the Cholesky manifold.

Inspired by the above observations, we leverage pullback techniques to introduce adaptive Riemannian metrics in this paper.
In particular, we first show that several Riemannian metrics on SPD manifolds, including LEM, LCM, and their generalizations, can be explained as pullback metrics from the standard Euclidean space.
We refer to these metrics as Pullback Euclidean Metrics (PEMs).
Then, we propose a general framework for characterizing the properties of PEMs. Our framework can explain the widely used LEM \cite{arsigny2005fast} and LCM \cite{lin2019riemannian}.
We focus on LEM on SPD manifolds and extend it into Adaptive Log-Euclidean Metrics (ALEMs).
Besides, we present a complete theoretical discussion on the properties of ALEMs. 
Compared with the existing Riemannian metrics, our metrics are adjustable, adapting to the characteristics of the datasets.
To the best of our knowledge, our work is the \textbf{first} to integrate learnable Riemannian metrics into Riemannian deep networks.
The effectiveness of our metrics is demonstrated by experiments as well as the applications to recently developed Riemannian building blocks, including Riemannian batch normalization \cite{chen2024liebn}, Riemannian residual blocks \cite{katsman2023riemannian}, and Riemannian classifiers \cite{nguyen2023building}.
Drawing on this, our \textbf{contributions} are summarized as follows:
\textbf{(a)} We reveal the connection of two popular Riemannian metrics (LEM and LCM) by the pullback technique and propose a general framework for PEMs;
\textbf{(b)} Based on our framework, we propose specific ALEMs on SPD manifolds and conduct comprehensive analyses in terms of the algebraic, analytic, and geometric properties;
\textbf{(c)} Extensive experiments on widely used SPD learning benchmarks demonstrate that our metrics exhibit consistent performance gain across datasets.

The rest of the paper is organized as follows: 
\cref{sec:preliminary} reviews some essential backgrounds of differential geometry and the geometry of SPD manifolds.
\cref{subsec:thk_lem_lcm} rethinks the existing LEM and LCM from the perspective of pullback metrics.
\cref{subsec:pem_spd} provides a detailed discussion on PEMs.
\cref{subsec:ada_rie_metric,app:subsec:differentials} extend the existing LEM into ALEMs based on the framework of PEMs.
\cref{sec:properties} extensively analyzes the geometric properties of ALEM.
\cref{sec:ada_param_layers} presents the application of our ALEM into SPD neural networks.
\cref{sec:param_learning} discusses the gradient computations and parameter updates involved in our methods.
\cref{sec:experiments} validates our metric on three datasets.
\cref{sec:app_other_riem_blocks} further applies our ALEM to re-design other Riemannian blocks.
\cref{sec:limitations} discusses the limitations of this work, and \cref{sec:conclusions} concludes this paper.
For better representation, all proofs are left in the supplement.

\section{Preliminaries} 
\label{sec:preliminary}
This section reviews some basic notations of differential geometry and the geometry of SPD manifolds.
For a more detailed review, please refer to the supplementary.

We first briefly review the idea of pullback, which is a common trick in geometry to study metrics.
\begin{definition} [Pullback Metrics] \label{def:pullback_metrics}
    Suppose $\calM,\calN$ are smooth manifolds, $g$ is a Riemannian metric on $\calN$, and $f:\calM \rightarrow \calN$ is smooth.
    Then the pullback of the tensor field $g$ by $f$ is defined point-wisely,
    \begin{equation} \label{eq:pullback_metrics}
        (f^*g)_p(V_1,V_2) = g_{f(p)}(f_{*,p}(V_1),f_{*,p}(V_2)),
    \end{equation}
    where $p \in \calM$, $f_{*,p}(\cdot)$ is the differential map of $f$ at $p$, and $V_i \in T_p\calM$.
    If $f^*g$ is positive definite, it is a Riemannian metric on $\calM$, which is called the pullback metric defined by $f$.
\end{definition}
The most common pullback metrics are the ones induced by diffeomorphism, \ie when $f$ is a diffeomorphism.

Next, we review the basic geometry of SPD manifolds.
We denote the set of $n \times n$ SPD matrices as $\spd{n}$, the set of $n \times n$ symmetric matrices as $\sym{n}$, and all the Cholesky matrices (lower triangular matrices with positive diagonal elements) as $\cho{n}$.
As shown in the previous literature \cite{arsigny2005fast,lin2019riemannian}, $\spd{n}$ and $\cho{n}$ form an SPD manifold and a Cholesky manifold, respectively.
For an SPD matrix $S$, the matrix logarithm $\mln(\cdot): \spd{n} \rightarrow \sym{n}$ is defined as
\begin{equation} \label{eq:mln} 
    \mln(S) = U \ln(\Sigma) U^\top,    
\end{equation}
where $S=U \Sigma U^\top$ is the eigendecomposition, and $\ln(\cdot)$ is the diagonal natural logarithm.

In \cite{arsigny2005fast}, LEM on $\spd{n}$ is introduced by Lie group translation. 
The standard LEM is further generalized into two-parameter families of $\orth{n}$-invariant metrics \cite{thanwerdas2023n}, namely $\biparamLEM$, by $\orth{n}$-invariant inner product on $\sym{n}$ 
\begin{equation}
    \langle X,X \rangle^{\biparam} = a \|X\|_\rmF + b \tr(X)^2, \forall X \in \sym{n},
\end{equation}
where $\|\cdot\|_\rmF$ is the Frobenius inner product, and $\biparam \in \bfst= \{\biparam \in \mathbb{R}^2 \mid \min (a, a+n b)>0\}$.
In \cite{lin2019riemannian}, LCM is derived on $\spd{n}$ from the Cholesky manifold $\cho{n}$ by Cholesky decomposition.
We denote $\biparamLEM$ and LCM as $\gbiparamlem$ and $\glcm$, respectively.
For an SPD matrix $P$ and a tangent vector $V$ in the tangent space $T_P\spd{n}$ at $P$, $\gbiparamlem$ is defined as
\begin{equation}
    \label{eq:biparamLEM}
    \gbiparamlem_{P} (V,V)=a \|\diffmln{P} (V)\|^2_{\rmF}+ b \tr(P^{-1}V)^2,
\end{equation}
where $\diffmln{P}$ is the differential map of matrix logarithm at $P \in \spd{n}$, $V$ is a tangent vector in the tangent space $T_P\spd{n}$ at $P$, $\biparam \in \bfst$.
Note that $\biparamLEM$ incorporates the standard LEM when $\biparam=(1,0)$.

For $L \in \cho{n}$ and $W \in T_L\cho{n}$, the metric on the Cholesky manifold \cite{lin2019riemannian} is defined as
\begin{equation} 
    \label{eq:cm}
    \gcm_L(W,W)=\sum_{i>j} W_{i j} W_{i j}+\sum_{j=1}^n W_{j j} W_{j j} L_{j j}^{-2},
\end{equation}
The LCM is the pullback metric by the Cholesky decomposition $\scrL$ from $\gcm$ \cite{lin2019riemannian}:
\begin{equation}
    \label{eq:lcm}
    \glcm = \scrL^{*}\gcm.
\end{equation}

\section{Adaptive Log-Euclidean Metrics}
\label{sec:ada_riem_metrics}
As mentioned in \cref{sec:intro}, pullbacks are ubiquitous for studying Riemannian metrics on SPD manifolds.
In this section, we further show that both $\biparamLEM$ and LCM are pullback metrics from the Euclidean space.
Inspired by this observation, we present a general framework for characterizing PEMs.
Then, we focus on generalizing LEM.

\subsection{Rethinking \texorpdfstring{$\biparamLEM$}{LEM} and LCM} \label{subsec:thk_lem_lcm}
Among the existing Riemannian metrics on the SPD manifold, LEM is popular in many applications, given its closed form for the Fréchet mean and clear vector space \& Lie group structures.
In addition, the nascent LCM, gaining increasing attention, also shares similar properties with LEM.
LEM is derived from the Lie group translation \cite{arsigny2005fast}, while LCM is derived by the pullback from $\cho{n}$ \cite{lin2019riemannian}.
Besides, $\biparamLEM$ is obtained by the pullback of LEM.
However, theoretically, the mathematical logic beneath their derivation can be the same. 
We denote $\tril{n}$ as the Euclidean space of $n \times n$ lower triangular matrices.
We define $\cln: \spd{n} \rightarrow \tril{n}$ as
\begin{equation} \label{eq:lcm_pullback_map}
    \cln(P)=\lfloor L \rfloor + \ln(\bbD(L)),
\end{equation}
where $L$ is the Cholesky factor of the SPD matrix $P$, $\lfloor L \rfloor$ is the strictly lower part of $L$, and $\bbD(L)$ is a diagonal matrix with diagonal elements of $L$. 
Then, we have the following theorem.
\begin{theorem} \label{thm:rethk_lem_lcm}
    $\biparamLEM$ is the pullback metric from the Euclidean space of $\sym{n}$ with an $\orth{n}$-invariant inner product $\langle , \rangle^{\biparam}$ by matrix logarithm.
    Specifically, the standard LEM is the pullback metric from the Euclidean space of $\sym{n}$ with the standard Frobenius inner product by matrix logarithm.
    LCM is the pullback metric from $\tril{n}$ with the Frobenius inner product by $\cln$.
\end{theorem}

As $n$-dimensional Euclidean spaces are naturally isometric, it can be directly obtained that both $\biparamLEM$ and LCM are pulled back from the standard Euclidean space $\sym{n}$.

\begin{corollary} \label{cor:biparamLEM_pem}
    $\biparamLEM$ and LCM are pullback metrics from $\sym{n}$ with standard Frobenius inner product.
\end{corollary}

\subsection{PEMs on SPD Manifolds}
\label{subsec:pem_spd}

In \cref{subsec:thk_lem_lcm}, we have shown how LEM is derived from matrix logarithm.
Besides, as shown in \cite{arsigny2005fast}, operations in Lie group and linear space on $\spd{n}$ are also induced from matrix logarithm.
Now, let us explain the underlying mechanism in detail.
A matrix logarithm is a diffeomorphism (a smooth bijection with a smooth inverse).
The property of bijection offers the possibility of transferring algebraic structures from $\sym{n}$ into $\spd{n}$.
The smoothness of matrix logarithm and its inverse suggest that smooth structures can be transferred into $\spd{n}$, like the Lie group and Riemannian metric.
More generally, given an arbitrary diffeomorphism $\phi:\spd{n} \rightarrow \sym{n}$, it suffices to pull various properties from the Euclidean space back to the SPD manifold $\spd{n}$ by $\phi$ as well.
Besides, the computation of the induced operators in $\spd{n}$ by $\phi$ is usually simple.


\begin{lemma} \label{lem:g_spd}
    Let $S_1, S_2, S \in \spd{n}, V_i \in T_S\spd{n}, k \in \bbRscalar$ and $g^{\rmE}$ be the Frobenius inner product in $\sym{n}$.
    $\phi:\spd{n} \rightarrow \sym{n}$ is a diffeomorphism, and $\phi_{*,S}$ is the differential at $S$.
    We define the following operations,
    \begin{align}
        \label{eq:phi_mul} 
        \text{Elements Addition: }& S_1 \phiMul S_2 = \phiinv( \phi(S_1)+\phi(S_2)),\\
        \label{eq:phi_sca_mul}
        \text{Scalar Product: }& k \phiMulScalar S_2 = \phiinv( k\phi(S_2)),\\
        \label{eq:phi_innerpro}
        \text{Inner Product: }& \langle S_1, S_2 \rangle_{\phi} = \langle \phi(S_1), \phi(S_2) \rangle,\\
        \label{eq:phi_g} 
        \text{Riemannian Metric: }& \gphi = \phi^*\geuc,
    \end{align}
    Then, we have the following conclusions:
    \begin{enumerate}
        \item \label{itm:spd_hilbet}
        $\{\spd{n}, \phiMul,\phiMulScalar, \langle \cdot, \cdot \rangle_{\phi} \}$ is a Hilbert space over $\bbRscalar$.
        \item 
        $\{\spd{n}, \phiMul \}$ is an Abelian Lie group.
        $\{\spd{n}, \gphi \}$ is a Riemannian manifold.
        The associated Riemannian operators are as follows
        \begin{align}
            \label{eq:dist_phi_spd}
            \dphi (S_1, S_2 ) &= \| \phi(S_1) - \phi(S_2) \|_\rmF,\\
            \label{eq:gene_rie_exp_spd}
            \rieexp_{S_1} V &= \phiinv(\phi(S_1)+\diffphi{S_1}V),\\
            \label{eq:gene_rie_log_spd}
            \rielog_{S_1}S_2 &= \diffphiinv{\phi(S_1)}(\phi(S_2)-\phi(S_1)),\\
            \label{eq:gene_pt_spd} \pt{S_1}{S_2}(V) &= \diffphiinv{\phi(S_2)} \circ \diffphi{S_1}(V),
        \end{align}
        where $\|\cdot\|_\rmF$ is the Frobenius norm, $V \in T_{S_1}\spd{n}$ is a tangent vector, $\rieexp_{S_1}$, $\rielog_{S_1}$ and $\pt{S_1}{S_2}$ are Riemannian exponential map at $S_1$, logarithmic map at $S_1$ and parallel transportation along the geodesics connecting $S_1$ and $S_2$ respectively, and $\phiinv_{*}$ is the differential maps of $\phiinv$.
        Then $\gphi$ is a bi-invariant metric, named Pullback Euclidean Metric (PEM) by $\phi$.
        \item
        $\phi$ is an isomorphism: (a) a linear isomorphism preserving the inner product; (b) a Lie group isomorphism; (3) a Riemannian isometry.
    \end{enumerate}
\end{lemma}
In fact, $\biparamLEM$ and LCM are special cases of \cref{lem:g_spd}, and so do linear space \& Lie group in \cite{arsigny2005fast} and Lie group in \cite{lin2019riemannian}.
In addition, neither \cite{arsigny2005fast} nor \cite{lin2019riemannian} reveals the Hilbert space structures in $\spd{n}$.
\subsection{Adaptive Log-Euclidean Metrics} \label{subsec:ada_rie_metric}
The key of \cref{lem:g_spd} lies in the diffeomorphism $\phi$.
If we have a proper $\phi$, Riemannian metrics on SPD manifolds can be induced.
In the following, we will present our mappings and then discuss the induced metrics.

As an eigenvalues function, the matrix logarithm in \cref{eq:mln} is reduced into a scalar logarithm, which is a diffeomorphism between $\bbRplus$ and $\bbRscalar$.
Following this hint, the eigenvalues-based diffeomorphism between $\spd{n}$ and $\sym{n}$ is reduced to scalar diffeomorphism between $\bbRplus$ and $\bbRscalar$. 
A very natural idea is to substitute the natural logarithm with scalar logarithms with arbitrary proper bases.
In particular, we can define a general diagonal logarithm $\log(\cdot)$ as
\begin{equation} \label{eq:diag_glog}
    \log_\alpha(X) = \diag(\log_{a_1}^{x_{11}},\log_{a_2}^{x_{22}},\cdots,\log_{a_n}^{x_{nn}}),
\end{equation}
where $\alpha = (a_1, a_2, \cdots, a_n) \in \bbRplus^{n} \setminus \{(1,1,\cdots,1)\}$ is the base vector, $\diag(\cdot)$ is the diagonalization operator, and $X$ is an $n \times n$ diagonal matrix.
By abuse of notation, we denote $\log_\alpha(\cdot)$ as $\log(\cdot)$ for a general diagonal logarithm, and $\log_a^{(\cdot)}$ as $\log^{(\cdot)}$ for a general scalar logarithm.
Specially, $a_1 = \cdots = a_n=e \Rightarrow \log(\cdot) = \ln(\cdot)$.
Together with eigendecomposition, a general matrix logarithm is:
\begin{equation} \label{eq:mlog}
     \mlog(S) = U \log_\alpha(\Sigma) U^\top,    
\end{equation}
where $S = U \Sigma U^\top$ is the eigendecomposition.
As a special case, when $\alpha=(e,e,\cdots,e)$, $\mlog = \mln$.
Similar to the scalar logarithm, we have the following proposition.
\begin{proposition}[Diffeomorphism] \label{props:diffeo_mlog}
    $\mlog$ is a diffeomorphism, a smooth bijection with a smooth inverse $\mlog^{-1}(\cdot):\sym{n} \rightarrow \spd{n}$ defined as
    \begin{equation}\label{eq:mgexp}
        \mlog^{-1}(X) = \mgexp(X) = U \balpha(\Sigma) U^\top,\\
    \end{equation}
    where $\balpha(\Sigma) = \diag(a_1^{\Sigma_{11}},a_2^{\Sigma_{22}},\cdots,a_n^{\Sigma_{nn}})$ is a diagonal exponentiation.
\end{proposition}
\begin{remark} \label{rmk:proposed_charts}
    Note that $\mlog(\cdot)$ should be more precisely understood as an arbitrary one from the following family
    \begin{equation}
        \{  \mlog^\alpha| \alpha = (a_1, \cdots, a_n) \in \bbRplus^{n} \setminus \{(1,\cdots,1)\}  \}.
    \end{equation}
    By abuse of notation, we will simply use $\mlog(\cdot)$.
    Besides, there could be some ambiguity in \cref{eq:mlog} under different arrangements of eigenvalues and eigenvectors.
    In fact, there is a correspondence between scalar $\log_{a_i}$ and eigenvalues \& eigenvectors.
    Please refer to Supp. B-A for more details.
\end{remark}
Since $\mlog$ is a diffeomorphism from $\spd{n}$ onto $\sym{n}$, all the results in \cref{lem:g_spd} hold true.
\begin{theorem} \label{thm:mlog_spd_properties}
    Following the notations in \cref{lem:g_spd}, we define $\mlogMul, \mlogMulScalar, \langle \cdot, \cdot \rangle_{mlog}$, and $g^{mlog}$ as \cref{eq:phi_mul}-\cref{eq:phi_g}.
    Then, we have the following conclusions:
    \begin{enumerate}
        \item \label{enum:hilbert}
        $\{\spd{n}, \mlogMul,\mlogMulScalar, \langle \cdot, \cdot \rangle_{mlog} \}$ is a Hilbert space over $\bbRscalar$.
        \item \label{enum:mlog_riem_spd}
        $\{\spd{n}, \mlogMul \}$ is an Abelian Lie group.
        $g^{mlog}$ is a Riemannian metric over $\spd{n}$.
        We call this metric Adaptive Log-Euclidean Metric (ALEM) and denote $g^{mlog}$ as $\galem$.
        The associated Riemannian operators are as follows
        \begin{align} 
            \label{eq:dist_mlog}
            &\dalem (S_1, S_2 ) = \| \mlog(S_1) - \mlog(S_2) \|_\rmF,\\
            \label{eq:rieexp_gmlog} 
            &\rieexp_{S_1} V = \mgexp(\mlog(S_1)+\diffmlog{S_1}V),\\
            \label{eq:rielog_gmlog} 
            &\rielog_{S_1}S_2 = \diffmgexp{X_1}(\mlog(S_2)-\mlog(S_1)),\\
            \label{eq:pt_mlog} 
            &\pt{S_1}{S_2}(V) = \diffmgexp{X_2} \circ \diffmlog{S_1}(V),
        \end{align}
        where $X_i = \mlog(S_i) \in \sym{n}$ for $i=1,2$.
        \item \label{enum:isomorphism}
        $\mlog$ is an isomorphism: (a) a linear isomorphism preserving the inner product; (b) a Lie group isomorphism; (3) a Riemannian isometry.
    \end{enumerate}
\end{theorem}
\begin{remark}
    Obviously, ALEM would vary with different $\mlog$.
    We thus use the plural to describe our metrics.
    Besides, our metrics could be learnable.
    This is why we call them adaptive metrics.
\end{remark}

Similar with $\biparamLEM$, we also can define $\biparamALEM$ as the pullback metric of $\orth{n}$-invariant inner product:
\begin{equation}
    \gbiparamalem= \mlog^*\gbiparamaE,
\end{equation}
where we denote the $\orth{n}$-invariant inner product $\langle , \rangle^{\biparam}$ as $\gbiparamaE$.
$\gbiparamalem$ also share the properties presented in \cref{thm:mlog_spd_properties}.
Nevertheless, this paper focuses on $\biparam=(1,0)$.

\subsection{Differentials of General Logarithms}
\label{app:subsec:differentials}
\cref{eq:rieexp_gmlog}-\cref{eq:pt_mlog} require the differential maps of $\mlog$ and $\mgexp$.
This subsection introduces the concrete formulae of the associated differential maps.
\begin{proposition}[Differentials] \label{props:diff_mgexp_mlog}
    For a tangent vector $V \in T_S\spd{n}$, the differential $\diffmlog{S} : T_S \spd{n} \rightarrow T_{\mlog(S)} \sym{n}$ of $\mlog$ at $S \in \spd{n}$ is given by
    \begin{equation}
        \diffmlog{S} (V) = Q+Q^\top + W,
    \end{equation}
    where $Q = D_U\log(\Sigma)U^\top$,
    \begin{align*}
        D_U &= (\begin{array}{ccc}
             (\sigma_1 I-S)^+ V u_1 & \cdots & (\sigma_n I-S)^+ V u_n
        \end{array}),\\
        W &= U \diag(\frac{u_1^\top V u_1}{\sigma_1 \ln{a_1}},\cdots,\frac{u_n^\top V u_n}{\sigma_n \ln{a_n}}) U^\top,        
    \end{align*}
    $()^+$ is the Moore–Penrose inverse, $u_1,\cdots,u_n$ are orthonormal eigenvectors of $S$, and the associated eigenvalues are $\sigma_1,\cdots,\sigma_n$.
    
    Symmetrically, for a tangent vector $\widetilde{V} \in T_X\sym{n}$, the differential $\diffmgexp{X} : T_X \sym{n} \rightarrow T_{\mgexp(X)} \spd{n}$ of $\mgexp$ at $X \in \sym{n}$ is given by
    \begin{equation} \label{eq:diff_mgexp}
        \diffmgexp{X} (\widetilde{V}) = \widetilde{Q}+\widetilde{Q}^\top + \widetilde{W},
    \end{equation}
    where $S = \widetilde{U} \widetilde{\Sigma}\widetilde{U}^\top$ is the eigendecomposition, $D_{\widetilde{U}}$ is defined similarly, $\widetilde{Q} = D_{\widetilde{U}}\balpha(\widetilde{\Sigma})\widetilde{U}^\top$, and
    \begin{equation*}
        \widetilde{W} = \widetilde{U} \diag(\ln^{a_1}a_1^{\widetilde{\sigma_1}}{\widetilde{u}_1^\top \widetilde{V} \widetilde{u}_1},\cdots,\ln^{a_n}a_n^{\widetilde{\sigma_n}}{\widetilde{u}_n^\top \widetilde{V} \widetilde{u}_n}) \widetilde{U}^\top.     
    \end{equation*}
\end{proposition}

In \cite{arsigny2005fast}, the differential of the matrix exponential is written as an infinite series.
The differential of our $\mgexp$ can also be rewritten in this way.
\begin{proposition}[Differential as Infinite Series] \label{props:diff_mgexp_series}
    Following the notation in \cref{props:diff_mgexp_mlog}, the differential of $\mgexp$ can also be formulated as
    \begin{equation} \label{eq:diff_mgexp_series}
        \begin{aligned}
            &\diffmgexp{X}(\widetilde{V}) \\
            &= \sum_{k=1}^{\infty} \frac{1}{k !}(\sum_{l=0}^{k-1} (\widetilde{P}X)^{k-l-1} (D_{\widetilde{P}}X+\widetilde{P}\widetilde{V}) (\widetilde{P}X)^l),
        \end{aligned}
    \end{equation}
    where $\widetilde{P}=\widetilde{U}B\widetilde{U}^\top$, $B=\diag(\ln^{a_1},\cdots,\ln^{a_n})$, $D_{\widetilde{P}}= D_{\widetilde{U}} B \widetilde{U}^\top + \widetilde{U} B D_{\widetilde{U}}^\top$.
\end{proposition}

When $\mgexp(\cdot)$ is reduced into matrix exponential, \cref{eq:diff_mgexp_series} becomes Eq. 8 in \cite{arsigny2005fast}, and our ALEM becomes exactly LEM.
\section{Properties OF ALEM}
\label{sec:properties}
Since our ALEMs are natural generalizations of LEM.
Therefore, intuitively, ALEMs would share every property of LEM.
This section introduces some useful properties of our ALEMs for machine learning, including Fréchet mean and invariance properties.

Fréchet means are important tools for SPD matrices learning \cite{harandi2018dimensionality,chakraborty2018statistical,brooks2019riemannian,chakraborty2020manifoldnorm}.
Like LEM, our ALEM also enjoys closed forms of Fréchet means.
We present a more general result, the weighted Fréchet mean.
\begin{proposition}[Weighted Fréchet Means] \label{props:geo_mean_spd}
    For $m$ points $S_1,\cdots S_m$ in SPD manifolds with associated weights $w_1,\cdots, w_m \in \bbRplus$,
    the weighted Fréchet mean $M$ over the metric space $\{\spd{n},\dalem\}$ has a closed form
    \begin{equation} \label{eq:fm_alem}
        M = \mgexp(\sum_{i=1}^{m} \frac{w_i}{\sum_{j=1}^{m} w_i}\mlog(S_i)).
    \end{equation}
\end{proposition}

Like LEM, although our ALEM does not conform with the affine-invariance, our ALEM enjoys some other kinds of invariance.
\begin{proposition}[Bi-invariance] \label{props:biinvariance}
    ALEM is a Lie group bi-invariant metric.
\end{proposition}
\begin{proposition} [Exponential Invariance] \label{props:exp_invariance}
    The Fréchet means under ALEM are exponential-invariant. 
    In other words, for $S_1,\cdots S_m \in \spd{n}$ and $\beta \in \bbRscalar$,
    \begin{equation}
        (\mathrm{FM}(S_1,\cdots S_m))^\beta = \mathrm{FM}(S_1^\beta,\cdots S_m^\beta),
    \end{equation}
    where $\mathrm{FM}(S_1,\cdots S_m))$ means the Fréchet mean of $S_1,\cdots S_m$.
\end{proposition}

Except for the exponential invariance, the Fréchet mean induced by our ALEM also satisfies various properties presented in \cite{ando2004geometric}.
\begin{proposition} \label{props:frechet_means_add_props}
    For any SPD matrices $A, B, C, A_0, B_0, C_0$, denote $\fm(A,B,C)$ as the Fréchet mean of $A,B,C$ under ALEM.
    Then the Fréchet mean satisfies the following properties.
    \begin{enumerate}[(U1)]
        \item \label{props:u1}
        Permutation invariance.
        For any permutation $\pi(\{ A,B,C \})$ of $\{ A, B, C\}$,  
        \item \label{props:u2}
        $\fm(A, A,A) = A$
    \end{enumerate}
    The following properties hold if $A, B, C, A_O, B_0, C_0$ commute.
    \begin{enumerate}[(V1)]
        \item  Joint homogeneity. \label{props:v1}
        $\fm(a A, b B, c C)=(a b c )^{1 / 3} \fm(A, B, C), \forall a,b,c >0$.
        
        \item Monotonicity.
        The map $(A, B, C) \mapsto \fm(A, B, C)$ is monotone, \ie, if $A \geq A_0$, $B \geq  B_0$, and $C \geq  C_0$, then $\fm(A,B,C) \geq  \fm(A_0,B_0,C_0)$ in the positive semidefinite ordering.
        \item Self-duality.
        $\fm(A, B, C)=\fm(A^{-1}, B^{-1}, C^{-1})^{-1}$.
        \item Determinant identity.  \label{props:v4}
        $\det \fm(A, B, C)=(\det A \cdot \det B \cdot \det C)^{1 / 3}$.
    \end{enumerate}
\end{proposition}

In fact, \cref{props:frechet_means_add_props} holds true for any finite number of SPD matrices. 
Besides, the geodesic distance induced by ALEMs has similarity invariance.

\begin{proposition}[Similarity Invariance] \label{props:sim_invariance}
    The geodesic distance under ALEM is similarity invariant.
    In other words, let $R \in SO(n)$ be a rotation matrix, $s \in \bbRplus$ is a scale factor.
    Given any two SPD matrices $S_1$ and $S_2$, we have
    \begin{equation}
        \dalem(S_1,S_2)=\dalem(s^2RS_1 R^\top,s^2RS_2 R^\top).
    \end{equation}
\end{proposition}
Let us explain a bit more about the above three kinds of invariance.
Firstly, among metrics on Lie groups, bi-invariant metrics are the most convenient ones \cite[Chapter V]{sternberg1999lectures}.
Secondly, exponential invariance offers a fast computation for Fréchet means under exponential scaling.
At last, similarity-invariance is significant for describing the frequently encountered covariance matrices \cite{arsigny2005fast}.

The above discussion focuses on theoretical side.
Now, let us reconsider \cref{eq:mlog} in a numerical way.

\begin{proposition} \label{prop:rewrit_general_log}
    $\mlog$ can be rewritten as
        \begin{align}
            \label{eq:rw_org_mlog} \mlog(S) 
            &= U \log_\alpha(\Sigma) U^\top,\\
            \label{eq:rw_mul_mlog}          
            &= U A \ln(\Sigma) U^\top, \\
            \label{eq:rw_div_mlog}          
            &= U \frac{\ln(\Sigma)}{B} U^\top,
        \end{align}
    where $\frac{X}{Y}$ is the diagonal division, $B=\diag(\ln^{a_1},\cdots,\ln^{a_n})$, and $A = \frac{I}{B}$.
\end{proposition}

Based on the above proposition, more analyses could be carried out from a numerical point of view.
First, $\mlog(\cdot)$ can balance the eigenvalues of an input SPD matrix $S$ by exploiting different bases for different eigenvalues.
In Riemannian algorithms, manifold-valued features usually contain vibrant information.
We expect that by the above adaptation, manifold-valued data could be better fitted and the learning ability of algorithms could be further promoted.
\begin{remark} 
Note that the discussion in \cref{subsec:ada_rie_metric} and \cref{sec:properties} can also be readily transferred into LCM, generating an adaptive version of LCM.
\end{remark}
\section{Applications to SPD Neural Networks} \label{sec:ada_param_layers}
Since Riemannian metrics are the foundations of Riemannian learning algorithms, our ALEM has the potential to rewrite Riemannian algorithms, especially the algorithms based on LEM.
Besides, the base vector in $\mlog$ could bring vibrant diversity to our ALEM.
This adaptive mechanism could help the algorithm better fit with complicated manifold-valued data. 
Especially in Riemannian neural networks, as we will show, optimization of base vectors can be easily embedded into the standard backpropagation (BP) process.
Therefore, we focus on the applications of our metrics to SPD neural networks.

In the existing SPD neural networks, on activation or classification layers, SPD features would interact with the logarithmic domain by matrix logarithm \cite{huang2017riemannian,zhen2019dilated, chakraborty2020manifoldnet,nguyen2021geomnet, chen2023riemannian}.
The underlying mechanism of this interaction is that the matrix logarithm is an isomorphism, identifying the SPD manifold under LEM with the Euclidean space $\sym{n}$.
This projection can, therefore, maintain the LEM-based geometry of SPD features.
However, in deep networks, the geometry might be more complex.
Since ALEM can vibrantly adapt to network learning, compared with the plain LEM, our ALEM could more faithfully respect the geometry of SPD deep features.
$\mlog$ thus possesses more advantages than the vanilla matrix logarithm $\mln$.
We, therefore, replace the vanilla matrix logarithm with our $\mlog$, to respect the more advantageous geometry, \ie the ALEM-based geometry.

We focus on the most classic SPD network, SPDNet \cite{huang2017riemannian}.
There are three basic layers in SPDNet, \ie BiMap, ReEig, and LogEig, which are defined as
\begin{align}
    \text{BiMap: }& S^{k} = W^k S^{k-1} W^k,\\
    \text{ReEig: }&S^{k}=U^{k-1} \max (\Sigma^{k-1}, \epsilon I_{n}) U^{k-1 \top},\\
    \text{LogEig: }&S^{k}=\mln(S^{k-1}),
\end{align}
where $W^k$ is semi-orthogonal and $S^{k-1}=U^{k-1} \Sigma^{k-1} U^{k-1 \top}$ is the eigendecomposition.
The BiMap (Bilinear Mapping) is a generalized version of conventional linear mapping.
The ReEig (Eigenvalue Rectification) mimics the ReLu-like nonlinear activation functions by eigen-rectification.
The LogEig layer projects SPD-valued data into the Euclidean space for further classification.

The matrix logarithm in the LogEig layer is substituted by our $\mlog$. 
We call this layer the adaptive logarithm (ALog) layer. 
We set the base vector $\alpha$ as a learnable parameter.
In this way, as $\mlog$ is an isomorphism, the network can implicitly respect the ALEM-based Riemannian geometry by learning the $\mlog$ explicitly.
Besides, since our ALog layer is independent of specific network architectures, it can also be plugged into other SPD deep networks.
\section{Parameters Learning} 
\label{sec:param_learning}

We first present the gradient computation and then discuss in detail how to optimize the parameters in the ALog layer.
\subsection{Gradients Computation} \label{subsec:gradients}

Two gradients need calculation in the proposed ALog layer: one w.r.t the parameters and another w.r.t the input of the ALog layer.
Since structural matrix decomposition is involved in $\mlog$, the following contents heavily rely on the structural matrix BP \cite{ionescu2015matrix}, the key idea of which is the invariance of first-order differential form.
For the ALog layer, it is essentially a special case of eigenvalue functions.
Based on the formula offered in \cite{bhatia2009positive} and matrix BP techniques presented in \cite{ionescu2015matrix}, we can obtain all the gradients, as presented in the following proposition.
\begin{proposition} \label{props:grad_mlog}
    Let us denote $X = \mlog(S)$, where $S \in \spd{d}$ is an input SPD matrix of the ALog layer.
    We have the following gradients:
    \begin{align}
        \label{eq:gradient_eigen_function} \nabla_{S} L
        &= U[K \odot(U^{T}(\nabla_{X} L) U)] U^{T},\\
        \nabla_{A} L   
        &= [U^\top (\nabla_{X} L) U] \odot \log(\Sigma),
    \end{align}
    where $S = U \Sigma U^\top$ is the eigendecomposition of an SPD matrix and matrix $K$ is defined as
    \begin{equation}
        K_{i j}= \begin{cases}\frac{f\left(\sigma_{i}\right)-f\left(\sigma_{j}\right)}{\sigma_{i}-\sigma_{j}} & \text { if } \sigma_{i} \neq \sigma_{j} \\ f^{\prime}\left(\sigma_{i}\right) & \text { otherwise }\end{cases}
    \end{equation}
    where $f(\sigma_i) = A_{ii}\log_e(\sigma_i)$ and $\Sigma=\diag(\sigma_1,\sigma_2,\cdots,\sigma_d$).
\end{proposition}

\subsection{Parameters Updates}
\begin{table*}[htbp]
    \small
    \centering
    \caption{Parameter Learning in the ALog Layer.}
    \label{tb:param_learning}
    \begin{tabular}{cccc}
    \toprule
    Name & Detail & Constraint & Method\\
    \midrule
    RELU & Optimizing base vector $\alpha$ (\cref{eq:rw_org_mlog}) & Positive & shift-ReLu $\max(\epsilon,\alpha)$\\
    MUL & Optimizing diagonal elements of $A$ (\cref{eq:rw_mul_mlog}) & Unconstrained & Standard BP\\
    DIV & Optimizing diagonal elements of $B$ (\cref{eq:rw_div_mlog}) & Unconstrained & Standard BP\\
    \bottomrule
    \end{tabular}
    \vspace{-2mm}
\end{table*}
Let us explain how to optimize the proposed layer in a standard backpropagation (BP) framework.
Denote the dimension of an input SPD matrix $S$ as $d \times d$.
Recalling \cref{eq:rw_org_mlog}-\cref{eq:rw_div_mlog}, there are three ways to implement parameter learning.
We could learn the base vector $\alpha$ in \cref{eq:rw_org_mlog}, diagonal matrix $A$ in \cref{eq:rw_mul_mlog}, or diagonal matrix $B$ in \cref{eq:rw_div_mlog}, respectively.

For learning $A$ in \cref{eq:rw_mul_mlog} or $B$ in \cref{eq:rw_div_mlog}, since the parameters (diagonal elements) lie in a Euclidean space $\bbR{d}$, the optimization can be easily integrated into the BP algorithm.
We call learning $A$ MUL and learning $B$ DIV.

For the case of learning $\alpha$ in \cref{eq:rw_org_mlog}, since $\alpha$ lies in a non-Euclidean space, specific updating strategies should be considered.
Without loss of generality, we focus on the case of a scalar parameter $a>0 \& a \neq 1$.
The condition of $a \neq 1$ can be further waived since we can set $a=1+\epsilon$ if $a=1$. 
Then, there is only one constraint about positivity.
We use the shift-ReLU of an unconstrained parameter, \ie $\max(\epsilon,a)$ with $\epsilon \in\bbRplus$.
This strategy is named RELU.
Other tricks like square are also feasible, but we will focus on the RELU.
In addition, positive scalar $a$ can be directly optimized by Riemannian optimization \cite{absil2009optimization}.
We further prove that this strategy completely equals learning $B$ directly.
For more details, please refer to the Supp. B-B.

Therefore, there are three ways of updates, \textit{i.e.,} RELU, DIV, and MUL, summarized in \cref{tb:param_learning}.

\section{Experiments} 
\label{sec:experiments}
In this section, we validate the efficacy of our approaches on multiple datasets.
We would like to clarify that our method does not necessarily aim to achieve the SOTA in a general sense for the following tasks but rather to promote the learning abilities of the family of SPD-based methods.

\subsection{Datasets and Settings}
As we discussed before, although the proposed ALog layers can be plugged into the existing SPD networks, we focus on the SPDNet framework \cite{huang2017riemannian}.
We follow the PyTorch code provided by SPDNetBN\footnote{https://proceedings.neurips.cc/paper/2019/file/\allowbreak 6e69ebbfad976d4637bb4b39de261bf7-Supplemental.zip} to reproduce SPDNet \& SPDNetBN and implement our approaches.

Following previous work \cite{huang2017riemannian,brooks2019riemannian}, we evaluate our methods on three datasets:
the HDM05 \cite{muller2007documentation} for skeleton-based actions recognition, the FPHA \cite{garcia2018first} for skeleton-based hand gestures recognition, and the AFEW \cite{dhall2018emotiw} for emotions recognition. 
The HDM05 dataset comprises motion capture data (MoCap) covering 130 action classes.
Each data point is a sequence of frames of 31 3D coordinates.
Each sequence can be represented by a $93 \times 93$ temporal covariance matrix.
For a fair comparison, we exploit the pre-processed $93 \times 93$ covariance features \footnote{https://www.dropbox.com/s/dfnlx2bnyh3kjwy/\allowbreak data.zip?dl=0} released by \cite{brooks2019riemannian}, which trims the dataset down to 2086 points scattered throughout 117 classes by removing some under-represented classes.
Following the settings in \cite{brooks2019riemannian},  we split the dataset into 50\% for training and 50\% for testing.
FPHA includes 1,175 clips of 45 different action categories.
Each frame is represented by 21 3D coordinates.
Similarly, each sequence can be modeled by a $63 \times 63$ covariance matrix.
For a fair comparison, we follow the experimental protocol in \cite{garcia2018first}, where 600 sequences are used for training, and 575 sequences are used for testing.
AFEW consists of 7 kinds of emotions, with 773 samples for training and 383 samples for validation.
We use the released pre-trained FAN\footnote{https://github.com/Open-Debin/Emotion-FAN} \cite{meng2019frame} to extract deep features and establish a $512 \times 512$ temporal covariance matrix for each video.

We denote $\{d_0, d_1,\cdots,d_L\}$ as the dimensions of each transformation layer in the SPDNet backbone.
Following the settings in \cite{brooks2019riemannian}, all networks are trained by the default Riemannian SGD \cite{becigneul2018riemannian} with a fixed learning rate $\gamma$ and batch size of 30.
To make ALog start from the vanilla matrix logarithm, the parameters in MUL, DIV, and RELU are initialized as 1,1 and $e$, respectively.
By abuse of notation, SPDNet-ALog-MUL is abbreviated as ALog-MUL, denoting that we substitute the LogEig layer (matrix logarithm) in SPDNet with our proposed ALog optimized by MUL.
All experiments use an Intel Core i9-7960X CPU with 32 GB RAM.

\subsection{Experimental Results} 
\label{sec:results_on_MLog}
\begin{table*}[htbp]
    \small
    \centering
    \caption{Results of ALog on the HDM05 Dataset.}
    \label{tb:mlog_on_HDM05}
    \begin{tabular}{c|ccc|ccc}
    \toprule
    Learning rate & \multicolumn{3}{c|}{$1e^{-2}$} & \multicolumn{3}{c}{$5e^{-2}$} \\
    \hline
    Architecture & \{ 93, 30\} & \{ 93, 70, 30\} & \{ 93, 70, 50, 30\} & \{ 93, 30\} & \{ 93, 70, 30\} & \{ 93, 70, 50, 30\} \\
    \hline
    SPDNet & 62.92±0.81 & 62.87±0.60 & 63.03±0.67 & 63.89±0.73 & 64.00±0.65 & 63.72±0.61 \\
    SPDNetBN & 63.03±0.75 & 58.27±1.7 & 52.02±2.34 & 63.75±0.69 & 48.78±5.15 & 37.84±6.10\\
    ALog-MUL & 63.52±0.75 & 63.86±0.58 & \textbf{63.94±0.44} & 64.4±0.68 & 64.60±0.69 & 64.36±0.49 \\
    ALog-DIV & \textbf{63.60±0.79} & 63.93±0.52 & 63.81±0.7 & \textbf{64.81±0.64} & \textbf{64.84±0.65} & \textbf{64.80±0.36} \\
    ALog-RELU & 63.02±0.79 & \textbf{63.94±0.64} & 63.14±0.65 & 63.97±0.75 & 64.10±0.63 & 63.78±0.46 \\
    \bottomrule
    \end{tabular}
    \vspace{-2mm}
\end{table*}

On the three datasets, the training epochs are set to be 200, 500, and 100.
We verify our ALog on the SPDNet with various architectures.
Besides, we further test the robustness of the proposed layer against different learning rates on the HDM05 and FPHA datasets.
Generally speaking, among all three kinds of implementation, \textbf{ALog-MUL} shows the most robust performance gain and achieves consistent improvement over the vanilla matrix logarithm. Besides, we could also observe that ALog-MUL is comparable to or even better than SPDNetBN, which yet brings much more complexity than our approach.
The main reason for the superiority of our ALog against the vanilla matrix logarithm is that our ALog can adaptively respect the vibrant geometry of SPD manifolds, depending on the characteristics of datasets, while only LEM can be respected by the matrix logarithm.
The following are detailed observations and analyses.

\textbf{Results on the HDM05 dataset.}
The 10-fold results are presented in \cref{tb:mlog_on_HDM05}, where dataset split and weights initialization are randomized.
Following \cite{huang2017riemannian}, three architectures are implemented on this dataset, \ie \{ 93, 30\}, \{ 93, 70, 30\}, and \{ 93, 70, 50, 30\}.
Generally speaking, endowed with the ALog, SPDNet would achieve consistent improvement.
Among all three kinds of implementation, RELU only brings limited improvement.
The reason might be that RELU fails to respect the innate geometry of the positive constraint.
There is another interesting observation worth mentioning.
In \cite{brooks2019riemannian}, only the result of SPDNetBN under the architecture of $\{93,30\}$ is reported on this dataset.
Our experiments show that with the network going deeper, SPDNetBN tends to collapse, while our ALog layer performs robustly in all settings.

\begin{figure}[htbp]
  \centering
  \includegraphics[trim={0mm 5mm 0mm 0mm}, width=\columnwidth]{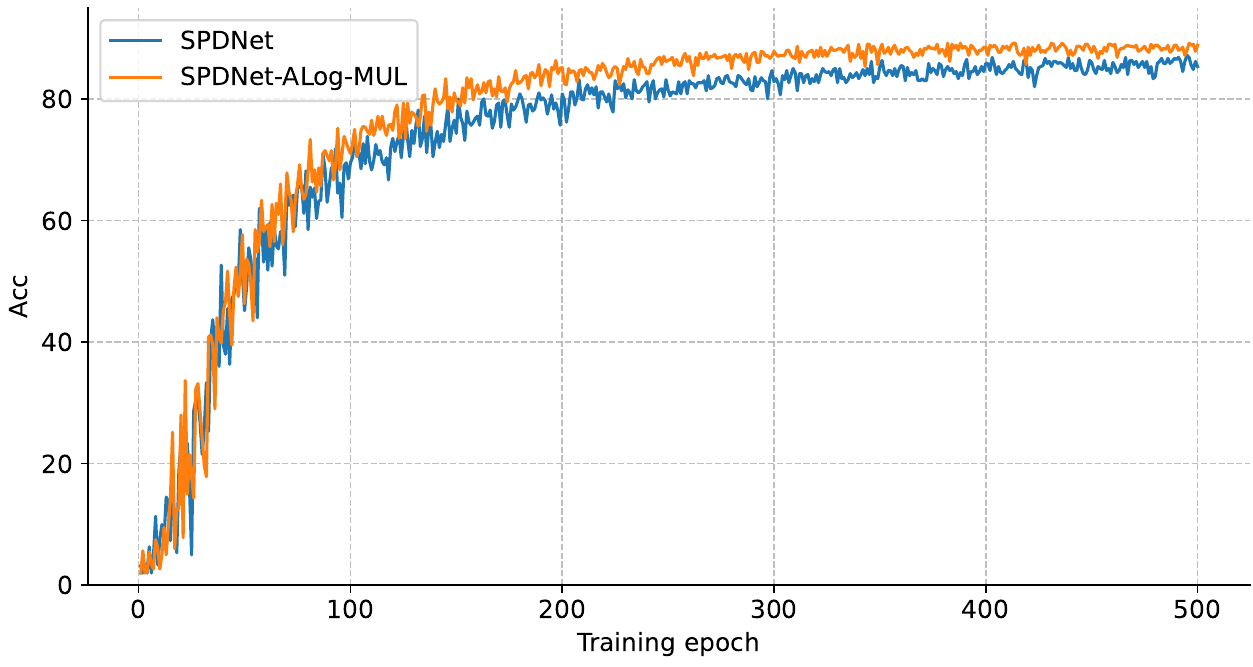}
  \caption{Accuracy Curves on the FPHA Dataset.}
  \label{fig:acc_fpha_mlog}
  \vspace{-2mm}
\end{figure}

\begin{table}[htbp]
    \small
    \centering
    \caption{Results of ALog on the FPHA Dataset.}
    \label{tb:mlog_on_FPHA}
    \resizebox{0.99\columnwidth}{!}{
    \begin{tabular}{ccccc}
        \toprule
        \multirow{2}[4]{*}{SPDNet} & \multirow{2}[4]{*}{SPDNetBN} & \multicolumn{3}{c}{ALog} \\
        \cmidrule{3-5}          &       & MUL   & DIV   & RELU \\
        \midrule
        85.73±0.80 & 86.83±0.74 & 87.8±0.71 & \textbf{88.07±1.13} & 86.65±0.68 \\
        \bottomrule
    \end{tabular}
    }
\end{table}

\textbf{Results on the FPHA dataset.}
We validate our approach on this dataset, with a learning rate of $1e^{-2}$, over 10-fold cross-validation on random initialization.
Since our experiments indicate that the vanilla SPDNet is already saturated with 1 BiMap layer, we just report the results on the architecture of $\{63,33\}$, which are presented in \cref{tb:mlog_on_FPHA}. 
Although DIV performs best on this dataset, it presents the biggest variance.
There is an underlying nonlinear scaling mechanism in the update of DIV, which might undermine its robustness.
Without loss of generality, let us focus on a single scalar parameter $b$ in \cref{eq:rw_div_mlog}.
The ultimate factor multiplied by the plain logarithm is $1/b$.
Therefore, the change of the multiplier after the update would be 
\begin{equation} \label{eq:nonlinear_div}
    1/(b-\Delta)-1/b=\Delta/[(b-\Delta)b].
\end{equation}
\cref{eq:nonlinear_div} will scale the original $\Delta$ to some extent.
This scaling mechanism might undermine the robustness of the ALog layer.
However, ALog-MUL achieves robust improvement and even surpasses SPDNetBN.
This again demonstrates the significance of our adaptive mechanism for Riemannian deep networks.
Finally, in terms of convergence analysis, accuracy curves with and without ALog are also reported in \cref{fig:acc_fpha_mlog}.

\begin{table}[htbp]
    \small
    \centering
    \caption{Results of ALog on the AFEW Dataset.}
     \label{tb:mlog_on_AFEW}
    \begin{tabular}{ccccc}
        \toprule
        Depth & 1 & 2 & 3 & 4 \\
        \midrule
        SPDNet & 48.53 & 46.89 & 48.24 & 47.22 \\
        SPDNetBN & 46.89 & 46.65 & 47.62 & 48.35 \\
        ALog-MUL & \textbf{48.57} & \textbf{48.13} & \textbf{49.45} & \textbf{50.62} \\
        ALog-DIV & 48.42 & 48.02 & 48.13 & 49.89 \\
        ALog-RELU & 48.06 & 47.25 & 48.86 & 48.1 \\
        \bottomrule
    \end{tabular}
\end{table}

\textbf{Results on the AFEW dataset.}
On this dataset, the learning rate is $5e^{-2}$ and we validate our method under four network architectures, \ie \{512, 100\}, \{512, 200, 100\}, \{512, 400, 200, 100\}, and \{512, 400, 300, 200, 100\}.
Note that, on this dataset, SPDNetBN tends to present relatively large fluctuations in performance, so we compute the median of the last ten epochs.
On various architectures, consistent improvement can be observed when SPDNet is endowed with our ALog.
In addition, MUL achieves the best among all three kinds of implementation.
Another interesting observation is that SPDNetBN seems ineffective on these deep features, while our methods show consistent superior performance, particularly obvious for our ALog-MUL. This indicates that our adaptive layer maintains effectiveness when applied to covariance matrices from deep features.

\textbf{Model complexity.}
Our ALog manifests the same complexity, no matter how it is optimized.
Without loss of generality, the discussion below focuses on ALog-MUL.
The extra computation and memory costs caused by the ALog layer are minor.
It only depends on the final dimension of the network.
Let us take the deepest one on the AFEW dataset as an example.
Our ALog only brings 100 unconstrained scalar parameters, while SPDNetBN needs an SPD matrix parameter for each Riemannian batch normalization (RBN) layer.
The total number of the parameters in RBN layers is $400^2+300^2+200^2$, which is much bigger than ours.
In addition, the SPDNetBN needs to store the running mean of SPD matrices in every RBN layer, while our ALog only needs to store a vector.
In terms of computation, the extra cost of our ALog is secondary as well.
The forward and backward computation of our ALog is generally the same as the plain matrix logarithm, while computation in the RBN layer is much more complex.
All in all, our ALog can consistently improve the performance of the SPDNet and achieve comparable or better results against SPDNetBN with much cheaper computation and memory costs.
\begin{figure}[htbp]
  \centering
  \includegraphics[width=\columnwidth]{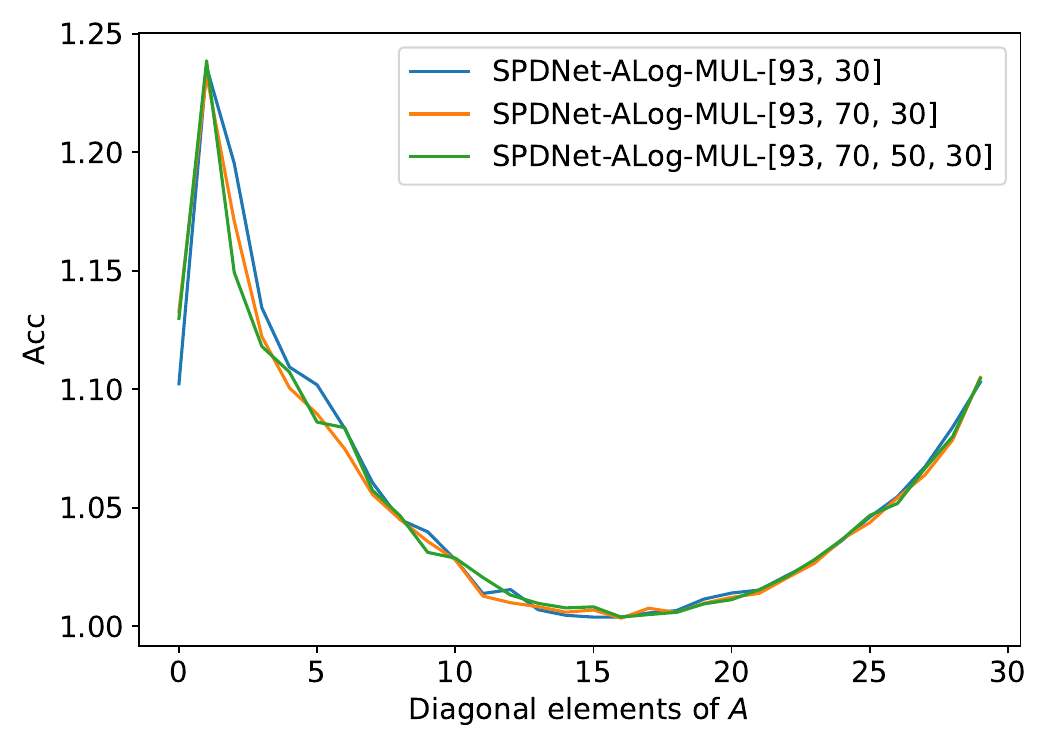}
  \caption{Visualization of Parameters in the ALog Layer on the HDM05 Dataset.}
  \label{fig:vis_params_hdm05}
\end{figure}

\begin{figure}[htbp]
  \centering
  \includegraphics[width=\columnwidth]{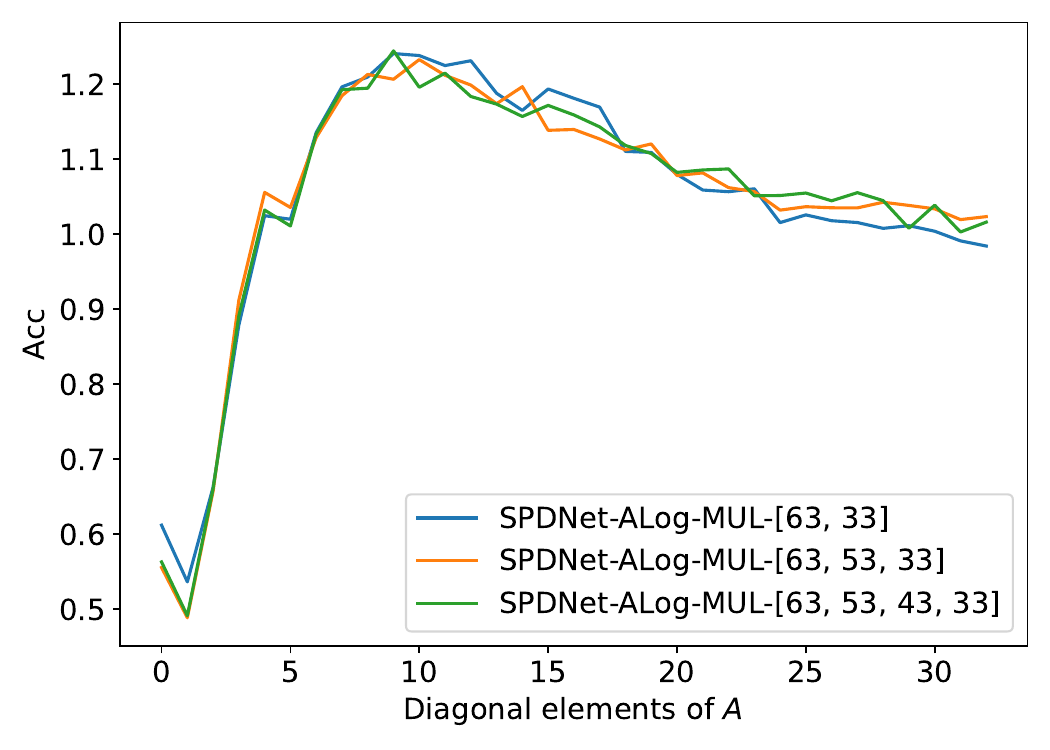}
  \caption{Visualization of Parameters in the ALog Layer on the FPHA Dataset.}
  \label{fig:vis_params_fpha}
\end{figure}


\textbf{Visualization.}
We visualize the final learned parameters of the ALog layer. 
Since ALog-MUL is the most robust strategy, we visualize the parameters of ALog-MUL. 
Specifically, we plot the final values of the diagonal elements of $A$ in \cref{eq:rw_mul_mlog} and visualize the results in \cref{fig:vis_params_hdm05,fig:vis_params_fpha}. 
We observe that the distribution of the parameters is consistent within the same dataset but varies between datasets. 
This indicates that our approach can capture vibrant patterns in different datasets, respecting their specific geometry.

\begin{table}[htbp]
    \small
    \centering
    \caption{Results of Fixed Bases on the HDM05 and FPHA Datasets.}
    \label{tab:fix_base_alog}
    \resizebox{0.99\columnwidth}{!}{
    \begin{tabular}{c|ccc|c}
        \toprule
        Dataset & \multicolumn{3}{c|}{HDM05} & FPHA \\
        Architecture & \{93, 30\} & \{93, 70, 30\} & \{93, 70, 50, 30\} & \{63, 33\} \\
        \hline
        SPDNet-Log2 & 63.93±0.81 & 63.54±0.50 & 63.98±0.63 & 86.65±0.67 \\
        SPDNet & 63.89±0.73 & 64.00±0.65 & 63.72±0.61 & 85.73±0.80 \\
        SPDNet-Log10 & 63.45±0.33 & 63.8±0.71 & 63.64±0.64 & 78.42±0.77 \\
        \rowcolor{gray!20}SPDNet-ALog-MUL & \textbf{64.4±0.68} & \textbf{64.60±0.69} & \textbf{64.36±0.49} & \textbf{87.8±0.71} \\
        \bottomrule
    \end{tabular}}
    \vspace{-2mm}
\end{table}

\textbf{Ablation studies.}
To further demonstrate the utility of the adaptive mechanisms in our approach, we further validate the ALog layer with fixed bases.
As decimal and binary are the two most common systems, we use $\log_{10}$ and $\log_{2}$ as examples of shrinking and expanding  $\log_e$. 
Specifically, we set $\log_\alpha$ = $\log_{10}$ and $\log_\alpha = \log_2$ in \cref{eq:rw_org_mlog}, respectively. 
We refer to the network with binary/decimal base as SPDNet-Log2/SPDNet-Log10. 
Note that when $\log_\alpha=\log_e$, \cref{eq:rw_org_mlog} is reduced to the vanilla matrix logarithm, and the network is our baseline, \ie SPDNet. 
We conduct 10-fold experiments on the HDM05 and FPHA datasets and set the learning rate to $5e^{-2}$ and $1e^{-2}$, respectively, while keeping the other settings consistent with previous experiments.
The results are presented in \cref{tab:fix_base_alog}. 
We observe that the fixed logarithms show similar or slightly worse results than the vanilla $\log_e$, while our ALog shows consistent improvement.
Besides, $\log_{10}$ does not converge in the FPHA dataset. 
In fact, $\log_{10}$ could shrink the gradient, slowing down convergence, especially under a small learning rate. 
In contrast, our ALog maintains consistent effectivity. 
In summary, our ALog can respect vibrant geometry induced by $\mlog$ and thus benefit SPD network learning.

\section{Applications to Other Riemannian Blocks}
\label{sec:app_other_riem_blocks}

Riemannian metrics are foundations for Riemannian neural networks.
Therefore, our ALEM can re-design basic blocks in Riemannian neural networks.
This section applies our ALEM to other Riemannian building blocks, including Riemannian batch normalization \cite{chen2024liebn}, Riemannian residual blocks \cite{katsman2023riemannian}, and Riemannian classifiers \cite{nguyen2023building}.
We also use the NTU60 \cite{shahroudy2016ntu} dataset as an example of the large-scale dataset. More implementation details are presented in Supp. C.

\subsection{Riemannian Batch Normalization}

\begin{table}[htbp]
  \centering
  \caption{Comparison of RBN methods on the HDM05 dataset.}
  \resizebox{\linewidth}{!}{
    \begin{tabular}{ccccc}
    \toprule
    Methods & Geometries & [93, 30] & [93, 70, 30] & [93, 70, 50, 30] \\
    \midrule
    None  & \na  & 63.89±0.73 & 64.00±0.65 & 63.72±0.61 \\
    SPDNetBN   & AIM   & 63.75±0.69 & 48.78±5.15 & 37.84±6.10 \\
    SPDBN & AIM   & 64.33±0.89 & 64.31±0.92 & 63.62±1.21 \\
    LieBN-LEM & LEM   & 63.67±0.85 & 65.77±0.89 & 65.34±0.83 \\
    \midrule
    LieBN-ALEM & ALEM  & \textbf{65.24±0.71} & \textbf{70.11±0.96} & \textbf{68.86±0.72} \\
    \bottomrule
    \end{tabular}
    }
  \label{tab:results_rbn}
\end{table}

In Euclidean neural networks, batch normalization \cite{ioffe2015batch} has been widely used since it can facilitate network training.
Recently, Chen~\etal~\cite{chen2024liebn} proposed a framework for Riemannian batch normalization (RBN) on Lie groups, referred to as LieBN.
LieBN can guarantee the normalization of sample statistics under the left- or right-invariant metric \cite[Prop. 4.2]{chen2024liebn}.
As shown in \cref{thm:mlog_spd_properties}, $\{\spd{n}, \mlogMul \}$ forms a Lie group.
Besides, \cref{props:biinvariance} demonstrates that ALEM is bi-invariant w.r.t. this group structure.
Therefore, LieBN under ALEM can also normalize Riemannian sample statistics.
We follow Alg. 1 and Thm 5.3 in \cite{chen2024liebn} to implement the LieBN under ALEM, denoted as LieBN-ALEM.
In addition, we compared LieBN-ALEM against other kinds of RBN methods, including AIM-based SPDNetBN \cite{brooks2019riemannian} and SPDBN \cite{kobler2022controlling}, and LieBN under LEM \cite{chen2024liebn} (LieBN-LEM).

Following previous work \cite{kobler2022controlling,brooks2019riemannian,chen2024liebn}, we adopt the SPDNet backbone. 
\cref{tab:results_rbn} presents the 10-fold average results on the HDM05 dataset under different network architectures.
Our LieBN-ALEM achieves the best performance compared with the other RBN methods.
Especially, the AIM-based SPDNetBN brings worse performance under deeper architectures.
In contrast, our LieBN-ALEM can consistently improve the performance across different architectures.
Besides, compared with LieBN-LEM, our LieBN-ALEM shows better performance, demonstrating the effectiveness of our ALEM.

\subsection{Riemannian Residual Blocks}

\begin{table}[htbp]
  \centering
  \caption{Experiments of RResNet under different geometries.}
    \begin{tabular}{ccc}
    \toprule
    Methods & HDM05 & NTU \\
    \midrule
    SPDNet & 63.89±0.73 & 45.90±1.11 \\
    RResNet-AIM & 63.82±0.58 & 45.22 ± 1.23 \\
    RResNet-LEM & 66.51±0.93 & 48.73±0.60 \\
    \midrule
    RResNet-ALEM & \textbf{69.03±1.06} & \textbf{57.09±0.59} \\
    \bottomrule
    \end{tabular}%
  \label{tab:results_rresnet_hdm05_ntu60}%
\end{table}%

ResNets \cite{he2016deep} have become ubiquitous in machine learning due to their beneficial learning properties.
Recently, Katsman~\etal~\cite{katsman2023riemannian} extended the Euclidean ResNet into Riemannian spaces, referred to as RResNet.
On the SPD manifold, the Riemannian residual block under a given metric $g$ is defined as
\begin{align}
    \label{eq:rresnet_eq1}
    g(S) &= \rieexp_{S}(\ell(S)),\\
    \ell(X) &= Q \diag \left(f(\spec (X))\right) Q^T,
\end{align}
where $\rieexp$ is the Riemannian exponentiation under $g$, $\ell: \spd{n} \rightarrow T\spd{n}$ constructs the vector field, $\spec(\cdot)$ is the spectral map that takes SPD matrices to a vector of their eigenvalues, $f: \bbR{n} \rightarrow \bbR{n}$ is parameterized as a neural network, and $Q \in \orth{n}$.
Since the Riemannian exponential in \cref{eq:rresnet_eq1} is metric-dependent, the Riemannian residual blocks vary under different metrics.
The Riemannian residual block under ALEM can be obtained by putting \cref{eq:rieexp_gmlog} into \cref{eq:rresnet_eq1}.
We need further to show the gradient w.r.t. $\mexp$.
As the inverse of \cref{eq:rw_mul_mlog}, $\mexp$ can be rewrote as
\begin{equation}
    \label{eq:mexp_rewrotten}
    \begin{aligned}
        \mexp(X) 
        &= U \balpha(\Sigma) U\top\\
        &= U \exp \left(\frac{\Sigma}{A} \right) U\top,
    \end{aligned}
\end{equation}
where $X=U \Sigma U^\top \in \sym{n}$ is the eigendecomposition.
Following \cref{props:grad_mlog}, we can obtain the backpropagation of $\mexp$, which is presented in the following.

\begin{proposition} \label{props:grad_mexp}
    Let us denote $X = \mexp(S)$ with $S \in \spd{d}$.
    We have the following gradients:
    \begin{align}
        \label{eq:gradient_mexp_wrt_S}
        \nabla_{S} L
        &= U[K \odot(U^{T}(\nabla_{X} L) U)] U^{T},\\
        \label{eq:gradient_mexp_wrt_A}
        \nabla_{A} L   
        &= [U^\top (\nabla_{X} L) U] \odot \left(\balpha(\Sigma) \frac{-\Sigma}{A^2} \right),
    \end{align}
    where $S = U \Sigma U^\top$ is the eigendecomposition of an SPD matrix and matrix $K$ is defined as
    \begin{equation}
        K_{i j}= \begin{cases}\frac{f\left(\sigma_{i}\right)-f\left(\sigma_{j}\right)}{\sigma_{i}-\sigma_{j}} & \text { if } \sigma_{i} \neq \sigma_{j} \\ f^{\prime}\left(\sigma_{i}\right) & \text { otherwise }\end{cases}
    \end{equation}
    where $f(\sigma_i) = e^{\frac{\sigma_i}{A_{ii}}}$ and $\Sigma=\diag(\sigma_1,\sigma_2,\cdots,\sigma_d$).
\end{proposition}

Following \cite{katsman2023riemannian}, we compare RResNet under different geometries on the HDM05 and NTU60 datasets.
\cref{tab:results_rresnet_hdm05_ntu60} reports the 10-fold and 5-fold average results on these datasets.
Compared with the vanilla SPDNet, RResNet-AIM brings little improvement, while LEM and ALEM show much better performance.
Especially, the ALEM-based RResNet can bring a clear performance improvement, underscoring the effectiveness of our ALEM.

\subsection{Riemannian Classifiers}

\begin{table}[htbp]
  \centering
  \caption{Comparison of Gyro MLRs on the NTU60 datasets.}
    \begin{tabular}{ccc}
    \toprule
    Learning Rates & $1e^{-2}$ & $5e^{-2}$ \\
    \midrule
    GyroMLR-AIM & 54.28±0.47 & 41.41±0.71 \\
    GyroMLR-LCM & 42.68±0.88 & 42.06±0.49 \\
    GyroMLR-LEM & 53.22±0.47 & 39.62±1.30 \\
    \midrule
    GyroMLR-ALEM & \textbf{56.21±0.39} & \textbf{51.65±0.44} \\
    \bottomrule
    \end{tabular}%
  \label{tab:gyro_mlr_ntu60}%
\end{table}%

Euclidean Multinomial Logistic Regression (MLR), which consists of FC and softmax, has become a standard classification block in Euclidean neural networks.
Inspired by this, Nguyen and Yang~\cite{nguyen2023building} extended the Euclidean MLR into the SPD manifolds by gyro structures \cite{nguyen2022gyro} for intrinsic classification, referred to as gyro MLR.
Three gyro MLRs under LCM, AIM, and LEM was introduced in \cite{nguyen2023building}.
Following the logic in \cite[Sec. 2.4.2]{nguyen2023building}, we can obtain the gyro MLR under ALEM.

\begin{theorem}[Gyro MLR] \label{thm:gyro_mlr_alem}
    Given an SPD feature $S \in \spd{n}$ and $C$ classes, the SPD gyro MLR under ALEM computes the multinomial probability of each class:
    \begin{equation}
    \label{eq:gyro_mlr_alem_start}
    \begin{aligned}
        &p(y=k \mid S)  \\
        &\propto
        \exp \left [ \langle \mlog(S)-\mlog(P_k), \diffmlog{P_k} (\tilde{A}_{k}) \rangle \right ],
    \end{aligned}
    \end{equation}
    where $k \in\{1, \ldots, C\}$, $P_k \in \spd{n}$, and $\tilde{A}_k \in T_{P_k} \spd{n}$.
\end{theorem}

Since $A_k$ lies in $T_{P_k} \spd{n}$, and $P_k$ varies during network training, $A_k$ cannot be viewed as a Euclidean parameter. Following \cite{ganea2018hyperbolic}, we set $\tilde{A}_{k}=\pt{I}{P_k}(A_k)$ with $A_k \in T_I \spd{n}$ (a fixed tangent space). Therefore, the RHS of \cref{eq:gyro_mlr_alem_start} becomes
\begin{equation}
        \exp \left [ \langle \mlog(S)-\mlog(P_k), \diffmlog{I} (A_k) \rangle \right ],
\end{equation}
As $\diffmlog{I} (A_k) \in T_0 \sym{n} \cong \sym{n}$, we view $\diffmlog{I} (A_k)$ as the parameter.

We use the SPDNet as the backbone. We compare Gyro MLR under our ALEM with the ones under LEM, LCM, and AIM on the NTU60 dataset.
\cref{tab:gyro_mlr_ntu60} presents the 5-fold average results under different learning rates.
Our ALEM outperforms the other metrics within the gyro MLR framework.
When the learning rate is $5e^{-2}$, our GyroMLR-ALEM shows more advantageous performance, especially compared with GyroMLR-LEM.
These results demonstrate that the Riemannian networks can benefit from the adaptivity of our ALEM.

\section{Limitations}
\label{sec:limitations}

Our approach presents a general framework for PEMs and specifically focuses on extending LEM. 
Despite the fast and simple computations of PEMs, there are several other types of Riemannian metrics on SPD manifolds, such as AIM \cite{pennec2006riemannian} and Bures-Wasserstein Metric (BWM) \cite{bhatia2019bures}.
These metrics do not belong to PEMs but have shown successful performance on different applications.
Therefore, the adaptive mechanisms of these types of Riemannian metrics should also be addressed in future work.

\section{Conclusion}
\label{sec:conclusions}

Riemannian metrics are foundations for Riemannian learning algorithms.
In this paper, we propose a general framework for characterizing PEMs on SPD manifolds.
According to this framework, we extend LEM into ALEMs for SPD matrix learning.
We also present comprehensive and rigorous theories of our metrics.
Extensive experiments indicate that SPD deep networks can benefit from our metrics.
\cref{eq:lcm_pullback_map} indicates that LCM is pulled back by Cholesky decomposition and diagonal logarithm.
Therefore, as a future avenue, the discussions in this paper can be readily transferred to LCM.

\bibliographystyle{IEEEtran}
\bibliography{Sections/ref}

\vsacle
\begin{IEEEbiographynophoto}{Ziheng Chen}
received the B.A. degree in logistics
management from Shandong University, Jinan, China, and M.S. degree in computer science and technology from Jiangnan University, Wuxi, China. He is currently working toward the Ph.D.
degree with the Multimedia and Human Understanding Group (MHUG), University of Trento, Trento, Italy. His research interests are machine learning, geometric deep learning, matrix manifolds, and matrix Lie groups.
\end{IEEEbiographynophoto}
\vsacle
\begin{IEEEbiographynophoto}{Yue Song}
received the B.Sc. cum
laude from KU Leuven, Belgium and the joint M.S.
summa cum laude from the University of Trento,
Italy and KTH Royal Institute of Technology, Sweden. He is currently working toward the Ph.D.
degree with the Multimedia and Human Understanding Group (MHUG), University of Trento, Trento,
Italy. His research interests are computer vision,
deep learning, and numerical analysis and
optimization.
\end{IEEEbiographynophoto}
\vsacle
\begin{IEEEbiographynophoto}{Tianyang Xu}
received the B.Sc. degree in electronic
science and engineering from Nanjing University,
Nanjing, China, in 2011. He received his Ph.D. degree
at the School of Artificial Intelligence and Computer Science, Jiangnan University, Wuxi, China, in
2019. He is currently an Associate Professor at the
School of Artificial Intelligence and Computer Science, Jiangnan University, Wuxi, China. His research
interests include visual tracking and deep learning.
\end{IEEEbiographynophoto}
\vsacle
\begin{IEEEbiographynophoto}{Zhiwu Huang}
received the B.Sc. degree in computer
science and technology from Huaqiao University,
Quanzhou, Fujian, China, in 2007, and the M.S.
degree in computer software and theory from
Xiamen University, Xiamen, Fujian, China, in 2010,
and the Ph.D. degree in computer science and technology from the Institute of Computing Technology
(ICT), Chinese Academy of Sciences (CAS),
Beijing, China, in 2015. He is currently a Lecturer affiliated with the Vision, Learning, and Control (VLC) group in the School of Electronics and Computer Science (ECS) at the University of Southampton.
His research interests include
computer vision, Riemannian computing, metric learning, and deep learning.
\end{IEEEbiographynophoto}
\vsacle
\begin{IEEEbiographynophoto}{Xiao-Jun Wu}
received the B.Sc. degree in mathematics from Nanjing Normal University, Nanjing,
China, in 1991. He received the M.S. degree and the
Ph.D. degree in pattern recognition and intelligent
systems from Nanjing University of Science and
Technology, Nanjing, China, in 1996 and 2002, respectively. He is a Professor in artificial intelligence
and pattern recognition at the Jiangnan University,
Wuxi, China. His research interests include pattern
recognition, computer vision, fuzzy systems, neural
networks, and intelligent systems. He has won several
domestic and international awards because of his research achievements. He
served as an associate editor for several international journals. He is currently a
Fellow of IAPR and AAIA.
\end{IEEEbiographynophoto}
\vsacle
\begin{IEEEbiographynophoto}{Nicu Sebe}
is a professor
at the University of Trento, Italy, leading the
research in the areas of multimedia information
retrieval and human behavior understanding. He was the General Co-Chair of the IEEE FG Conference 2008 and ACM Multimedia 2013; the Program Chair of the International Conference on Image and Video Retrieval in 2007 and 2010, ACM Multimedia 2007 and 2011, and the ICCV 2017 and ECCV 2016; and the General Chair of ACM ICMR 2017. He is a fellow of the IAPR.
\end{IEEEbiographynophoto}

\renewcommand{\appendixname}{Supplementary Material}
\crefalias{section}{suplement}
\crefalias{subsection}{suplement}
\clearpage
\appendices


\section{Preliminaries}
\label{app:sec:preliminaries}
\subsection{Smooth Manifolds}
We first recap some basic definitions related to this work on smooth manifolds.
Please refer to \cite{loring2011introduction,lee2013smooth} for in-depth understanding. 

The most important properties of manifolds are locally Euclidean, which are described by coordinate systems.

\begin{definition}[Coordinate Systems, Charts, Parameterizations] \label{def:Parameterization}
A topological space $\calM$ is locally Euclidean of dimension $n$ if every point in $\calM$ has a neighborhood $U$ such that there is a homeomorphism $\phi$ from $U$ onto an open subset of $\mathbb{R}^{n}$. 
We call the pair $\{ U, \phi: U \rightarrow \mathbb{R}^{n}\}$ as a chart, $U$ as a coordinate neighborhood, the homeomorphism $\phi$ as a coordinate map or coordinate system on $U$, and $\phi^{-1}$ as a parameterization of $U$. 
\end{definition}

Intuitively, a coordinate system is a bijection that locally identifies the Euclidean space with the manifold.
It locally preserves the most basic properties in a manifold, the topology. 
Topological manifolds, which are foundations of smooth manifolds, can be defined.
\begin{definition}[Topological Manifolds]
    A topological manifold is a locally Euclidean, second countable, and Hausdorff topological space.
\end{definition}

Compatibility is further required in smooth manifolds to define smooth structures or operations.
\begin{definition}[$C^{\infty}$-compatible] \label{def:compatible}
Two charts $\{ U, \phi_1: U \rightarrow \mathbb{R}^{n} \},\{ V, \phi_2: V \rightarrow \mathbb{R}^{n} \}$ of a locally Euclidean space are $C^{\infty}$-compatible if the following two composite maps
\begin{equation}
    \begin{aligned}
        \phi_1 \circ \phi_2^{-1} &: \phi_2(U \cap V) \rightarrow \phi_1(U \cap V), \\
        \quad \phi_2 \circ \phi_1^{-1} &: \phi_1(U \cap V) \rightarrow \phi_2(U \cap V)        
    \end{aligned}
\end{equation}
are $\cinf$.
\end{definition}

By abuse of notation, we view $\phi$ alternatively as a chart or map according to the context, and abbreviate $C^{\infty}$-compatible as compatible.

\begin{definition}[Atlases] \label{def:atlas}
A $C^{\infty}$ atlas or simply an atlas on a locally Euclidean space $\calM$ is a collection $\calA=\{ \{ U_{\alpha}, \phi_{\alpha} \} \}$ of pairwise $\cinf$-compatible charts that cover $\calM$.
\end{definition}

An atlas $\mathcal{A}$ on a locally Euclidean space is said to be maximal if it is not contained in a larger atlas. 
With a maximal atlas, smooth manifold can be defined.
\begin{definition}[Smooth Manifolds] 
A smooth manifold is defined as a topological manifold endowed with a maximal atlas.
\end{definition}

We call the maximal atlas of a smooth manifold its differential structure.
In addition, every atlas $\calA$ is contained in a unique maximal atlas $\calA^+$ \cite{loring2011introduction}.
Therefore, an atlas can be used to identify the differential structure of a smooth manifold.
In this paper, manifolds always mean smooth manifolds.
Now, we can define the smoothness of a map between manifolds.
\begin{definition}[Smoothness] \label{def:smoothness}
Let $\calN$ and $\calM$ be smooth manifolds, and $f: \calN \rightarrow \calM$ a continuous map, $f(\cdot)$ is said to be $\cinf$ or smooth, if there are atlases $\calA_n$ for $\calN$ and $\calA_m$ for $\calM$ such that for every chart $\{ U, \phi \}$ in $\calA_n$ and $\{ V, \psi \}$ in $\calA_m$, the map
\begin{equation}
    \psi \circ F \circ \phi^{-1}: \phi\left(U \cap f^{-1}(V)\right) \rightarrow \mathbb{R}^{m}
\end{equation}
is $C^{\infty}$.
\end{definition}
In elementary calculus, smooth functions have derivatives.
In manifolds, derivatives are generalized into differential maps.
\begin{definition} [Differential Maps]
    Let $f: \calN \rightarrow \calM$ be a $C^{\infty}$ map between two manifolds. At each point $p \in \calN$, the map $f$ induces a linear map of tangent spaces, called its differential at $p$,
    \begin{equation}
        f_{*,p}: T_p \calN \rightarrow T_{f(p)} \calM.
    \end{equation}
    $f_{*,p}$ can be locally represented by the Jacobian matrix under a chart $\{ U, \phi \}$ about $p$ and a chart $\{ V, \psi \}$ about $f(p)$,
    \begin{equation}
        f_{*,p} := \frac{\partial f}{\partial x} := \frac{\partial \psi f \phi^{-1}}{\partial x},
    \end{equation}
    where $\frac{\partial f}{\partial x}$ is called the derivative (Jacobian matrix) of $f$ under the charts of $\{ U, \phi \}$ and $\{ V, \psi \}$.
\end{definition}
With the definition of smoothness, it is possible to define smooth algebraic structures on a manifold, \ie Lie groups.
Intuitively, a Lie group is an integration of algebra (group) and geometry (manifold).
\begin{definition}[Lie Groups] \label{def:lie_group}
A manifold is a Lie group, if it forms a group with a group operation $\odot$ such that $m(x,y) \mapsto x \odot y$ and $i(x) \mapsto x_{\odot}^{-1}$ are both smooth, where $x_{\odot}^{-1}$ is the group inverse of $x$.
\end{definition}
\subsection{Riemannian Manifolds}
When manifolds are endowed with Riemannian metrics, various Euclidean operators can find their counterparts in manifolds.
A plethora of discussions can be found in \cite{do1992riemannian}.

\begin{definition}[Riemannian Manifolds] \label{def:riem_manifold}
A Riemannian metric on $\calM$ is a smooth symmetric covariant 2-tensor field on $\calM$, which is positive definite at every point.
A Riemannian manifold is a pair $\{\calM,g\}$, where $\calM$ is a smooth manifold and $g$ is a Riemannian metric.
\end{definition}
As a basic fact in differential geometry, every smooth manifold is a Riemannian manifold \cite[Prop.~2.10]{do1992riemannian}.
Therefore, in the following, we will alternatively use manifolds or Riemannian manifolds.
\begin{definition} [Pullback Metrics] \label{def:pullback_metrics_app}
    Suppose $\calM,\calN$ are smooth manifolds, $g$ is a Riemannian metric on $\calN$, and $f:\calM \rightarrow \calN$ is smooth.
    Then the pullback of a tensor field $g$ by $f$ is defined point-wisely,
    \begin{equation} \label{eq:pullback_metrics_app}
        (f^*g)_p(V_1,V_2) = g_{f(p)}(f_{*,p}(V_1),f_{*,p}(V_2)),
    \end{equation}
    where $p$ is an arbitrary point in $\calM$, $f_{*,p}(\cdot)$ is the differential map of $f$ at $p$, and $V_1,V_2$ are tangent vectors in $T_p\calM$.
    If $f^*g$ is positive definite, it is a Riemannian metric on $\calM$, called the pullback metric defined by $f$.
\end{definition}

\begin{definition}[Isometries] \label{def:isometry}
If $\{M, g\}$ and $\{\widetilde{M}, \widetilde{g}\}$ are both Riemannian manifolds, a smooth map $f: M \rightarrow$ $\widetilde{M}$ is called a (Riemannian) isometry if it is a diffeomorphism that satisfies $f^{*} \tilde{g}=g$.
\end{definition}
If two manifolds are isometric, they can be viewed as equivalent.
Riemannian operators in these two manifolds are closely related. 
\begin{definition}[Bi-invariance] \label{def:bi_invariance}
A Riemannian metric $g$ over a Lie group $\{G, \odot\}$ is left-invariant, if for any $x,y \in G$ and $V_1,V_2 \in T_x\calM$, 
\begin{equation}
    g_y(V_1,V_2) = g_{L_x(y)}(L_{x*,y}(V_1), L_{x*,y}(V_2)),
\end{equation}
where $L_x(y) = x \odot y$ is left translation, and $L_{x*,y}$ is the differential map of $L_x$ at $y$.
Right-invariance is defined similarly.
A metric over a Lie group is bi-invariant if both left- and right-invariant. 
\end{definition}
Bi-invariant metrics are the most convenient metrics on Lie, as they enjoy many excellent properties \cite[Ch.~V]{sternberg1999lectures}. 

The exponential \& logarithmic maps and parallel transportation are also crucial for Riemannian approaches in machine learning.
To bypass the notation burdens caused by their definitions, we review the geometric reinterpretation of these operators \cite{pennec2006riemannian, do1992riemannian}.
In detail, in a manifold $\calM$, geodesics correspond to straight lines in the Euclidean space.
A tangent vector $\overrightarrow{x y} \in T_x\calM$ can be locally identified to a point $y$ on the manifold by geodesic starting at $x$ with initial velocity of $\overrightarrow{x y}$, i.e. $y=\rieexp_x(\overrightarrow{x y})$.
On the other hand, the logarithmic map is the inverse of the exponential map, generating the initial velocity of the geodesic connecting $x$ and $y$, i.e. $\overrightarrow{x y}=\rielog_x(y)$.
These two operators generalize the idea of addition and subtraction in Euclidean space.
For the parallel transportation $\pt{x}{y}(V)$, it is a generalization of parallelly moving a vector along a curve in Euclidean space.
we summarize the reinterpretation in \cref{tb:reinter_riem_operators}.
\begin{table}
    \centering
    \caption{Reinterpretation of Riemannian Operators.}
    \label{tb:reinter_riem_operators}
    \begin{tabular}{ccc}
    \toprule
    Operations & Euclidean spaces & Riemannian manifolds \\
    \midrule
    Straight line & Straight line & Geodesic \\
    Subtraction & $\overrightarrow{x y}=y-x$ & $\overrightarrow{x y}=\log _x(y)$ \\
    Addition & $y=x+\overrightarrow{x y}$ & $y=\exp _x(\overrightarrow{x y})$ \\
    Parallelly moving & $V \rightarrow V$ & $\pt{x}{y}(V)$\\
    \bottomrule
    \end{tabular}
\end{table}

\subsection{LEM and LCM on the SPD Manifold} \label{subsec:geom_of_spd}
This subsection briefly reviews LEM \cite{arsigny2005fast} and LCM \cite{lin2019riemannian}.

Matrix logarithm $\mln(\cdot): \spd{n} \rightarrow \sym{n}$ and $\cln(\cdot): \spd{n} \rightarrow \tril{n} $ are defined as,
\begin{align}
     \mln(S) &= U \ln(\Sigma) U^\top,\\
     \cln(P)&=\clnchart(\scrL(S)),
\end{align}
where $S=U \Sigma U^\top$ is the eigendecomposition, $L = \scrL(S)$ is the Cholesky decomposition ($S=LL^\top$), $\clnchart(L) = \lfloor L \rfloor + \ln(\bbD(L))$ is a coordinate system from the $\cho{n}$ manifold onto the Euclidean space $\tril{n}$ \cite{lin2019riemannian}, $\lfloor L \rfloor$ is the strictly lower triangular part of $L$, $\bbD(L)$ is the diagonal elements, and $\ln(\cdot)$ is the diagonal natural logarithm.
We name $\cln$ as the Cholesky logarithm, since we will rely on it many times in the following proof.
Note that topologically, $\tril{n} \simeq \sym{n} \simeq \bbR{n(n+1)/2}$, since their metric topology all comes from the Euclidean metric tensor.
Based on matrix logarithm, \cite{arsigny2005fast} proposed LEM by Lie group translation, while based on Cholesky logarithm, \cite{lin2019riemannian} proposed LCM, by an isometry between $\spd{n}$ and $\cho{n}$.
In the main paper, we argued that LEM and LCM are basically the same, in the sense of high-level mathematical abstraction.

The Riemannian metric and associated geodesic distance under the LEM are defined by:
\begin{align}
    \label{eq:metric_lem} \glem_{S}(V_1,V_2) &= \geuc( {\mln}_{*,S} ( V_{1}), {\mln}_{*,S} ( V_{2}) ),\\
    \dlem(S_1, S_2) &= \| \mln(S_1) - \mln(S_2)\|_\rmF,
\end{align}
where $S \in \spd{n}$, $V_1, V_2 \in T_S\spd{n}$ are tangent vectors, ${\mln}_{*,S}(\cdot)$ is the differential map of matrix logarithm at $S$, $\geuc$ is the standard Euclidean metric tensor, and $\| \cdot \|_F$ is Frobenius norm.
Note that since $\geuc$ is the same at every point, we simply omit the subscript.
Besides, element-wise and scalar multiplication are also induced by $\mln$:
\begin{align}
    S_1 \mlnMul S_2 & = \mexp(\mln(S_1) + \mln(S_2)),\\
    \lambda \mlnMulScalar S & = \mexp(\lambda \mln(S)),
\end{align}
where $\mexp(X) = U \exp(\Sigma) U^\top$ is the matrix exponential.
As is proven in \cite{arsigny2005fast}, $\{ \spd{n}, \mlnMul\}$ and $\{ \spd{n}, \mlnMul,\mlnMulScalar\}$ form a Lie group and vector space, respectively.
Besides, the metric $\glem$ defined on Lie group $\{ \spd{n}, \mlnMul\}$ is bi-invariant.

The Riemannian metric and geodesic distance under LCM is
\begin{align}
     \label{eq:metric_lcm} 
    \glcm_S(V_{1}, V_{2} ) &= \gcm_L(L(L^{-1} V_1 L^{-\top})_{\frac{1}{2}},L(L^{-1}V_2L^{-\top})_{\frac{1}{2}}),\\
    \dlcm(S_1, S_2) &= \{\|\lfloor L_1\rfloor-\lfloor L_2\rfloor\|_{\rmF}^{2}\\
    &+\|\ln( \bbD(L_1))-\ln (\bbD(L_2))\|_{\rmF}^{2}\}^{\frac{1}{2}}, 
\end{align}
where $S \in \spd{n}$, $V_1,V_2 \in T_S\spd{n}$, $X_{\frac{1}{2}} = \lfloor X \rfloor + \bbD(X)/2$, and $\gcm_L(\cdot, \cdot)$ is the Riemannian metric on $\cho{n}$, defined as
\begin{align} \label{eq:metric_cholesky}
    \gcm_{L}(X, Y) 
    &= \geuc (\lfloor X\rfloor,\lfloor Y\rfloor)\\ 
    &+ \geuc(\bbD(L)^{-1} \bbD(X), \bbD(L)^{-1} \bbD(Y)).
\end{align}
The group operation in \cite{lin2019riemannian} is defined as follows:
\begin{equation}
    S_1 \clnMul S_2 = \scrL^{-1} (\lfloor L_1 \rfloor + \lfloor L_2 \rfloor + \bbD(L_1)\bbD(L_2)),
\end{equation}
where $\scrL^{-1}(\cdot)$ is the inverse map of Cholesky decomposition.
$\{\spd{n}, \clnMul\}$ is proven to be a Lie group \cite{lin2019riemannian}.
Similar to LEM, $\glcm$ is bi-invariant.

\section{Additional Discussions on the ALEM}
In this section, we present additional discussions on our ALEM.
All the proofs are placed in \cref{app:proofs}.
\subsection{Well-definedness of General Matrix Logarithm}
\label{app:subsec:well_difined_glog}

In \cref{eq:mlog}, due to the page limit, we did not clarify specific correspondence between eigenvalue and diagonal logarithm.
Here, we present detailed clarification.
Note that in implementation, like PyTorch or Matlab, this is no need to worry about this issue, as the outputs of eigendecomposition are always ordered.

We rewrite the eigendecomposition as $S = \sum \sigma_i E_i$ where $E_i=u_i u_i^\top$and $u_i$ is the corresponding eigenvector in $U$. Let $S$ be an $n \times n $ SPD matrix and $P_n$ be a set of all permutations of $\{n,\cdots, 1\}$, known as a permutation group. Changing the order of $\{n,\cdots, 1\}$ can be viewed as a permutation, so we use $\pi \in P_n$ to represent the corresponding changed order.

Assume the eigenvalues $\sigma_i$ are sorted in ascending order, \emph{i.e.,} $\sigma_1 \leq \cdots, \leq \sigma_n$. To clarify the definition of "the $i$-th eigenvalues", we refer to the $i$-th eigenvalue to the $i$-th pair from the ordered eigenpair sequence $(\sigma_1,u_1),\cdots,(\sigma_n,u_n)$. Since each eigenvector $u_{i}$ is unique, it is safe to say the eigenvalues are ordered, and the $i$-th eigenvalue/eigenvector pair is unique.

Let $\log_\alpha(\Sigma)$ denotes imposing scalar logarithm $\log_{a_i}$ to the $i$-th eigenvalue $\sigma_i$. Then $\phi_{m l o g}$ is rewritten as $\phi_{m l o g}(S)=\sum \log _{a_i}^{\sigma_i} E_i$, where $S = \sum \sigma_i E_i$. In this way, $\phi_{m l o g}$  is clearly well-defined. By definition, we can observe that the output of $\phi_{m l o g}$ does not depend on the order in eigendecomposition.

Suppose there are two eigendecomposition with different orders, \emph{i.e.,} $S= U\Sigma U^\top = \tilde{U} \tilde{\Sigma}\tilde U^\top$ where $\tilde{U}, \tilde\Sigma$ are the rearrangement of $U,\Sigma$. There exists a $\pi \in P_n$  such that for each $j$, there is a unique $i$, satisfying $\tilde u _j = u_{\pi(i)}$ and $\tilde\sigma_j=\sigma_{(i)}$. We then have $\sum \log _{a_i}^{\sigma_i} E_i$  for $S= U\Sigma U^\top$ and $\sum \log_{a_{\pi(i)}}^{\sigma_{\pi(i)}} E_{\pi(i)}$ for $S=\tilde U\tilde{\Sigma}\tilde U^\top$, which indicates the two eigendecomposition are equivalent. 

\subsection{Learning Base Vectors by Riemannian Optimization} \label{app:subsec:geom}

We focus on a single element $a$ of $\alpha$ in \cref{eq:rw_org_mlog}.
As discussed in the main paper, $a$ satisfying $a>0 \& a \neq 1$.
The condition of $a \neq 1$ can be further waived since we can set $a=1+\epsilon$ if $a=1$. Then, there is only one constraint about positivity.
A geometric way to deal with positivity is to view $a$ as a point in a 1-dimensional SPD manifold.
We call this strategy GEOM.
Then, we have the following updating formula for GEOM.
\begin{proposition} \label{propos:param_by_geom}
    Viewing a positive scalar $a$ as a point in a 1-dimensional SPD manifold, we have the following updating formula for Riemannian stochastic gradient descent (RSGD).
    \begin{equation} \label{eq:update_pos_scalar}
        a^{(t+1)} = a^{(t)}e^{-\gamma^{(t)} a^{(t)}  \nabla_{a^{(t)}} L},
    \end{equation}
    where $\nabla_{a^{(t)}} L$ is the Euclidean gradient of $a$ at $a^{(t)}$, $\gamma^{(t)}$ is the learning rate, and $e^{(\cdot)}$ is the natural exponentiation.
\end{proposition}

Besides, by \cref{eq:update_pos_scalar}, we could prove that GEOM is equivalent to DIV, which is given in the following proposition.
\begin{proposition} \label{props: geom_equivalent_div}
    For parameters learning in $\mlog$, optimizing the base vector $\alpha$ by RSGD is equivalent to optimizing the divisor matrix $B$ by Euclidean stochastic gradient descent (ESGD).
\end{proposition}

\section{Implementation Details of Additional Applications}
\label{sub:sec:imp_details}

\subsection{Details on the NTU60 Dataset}
\textbf{NTU60} \cite{shahroudy2016ntu}. It has 56,880 sequences of 3D skeleton data classified into 60 classes, where each frame contains the 3D coordinates of 25 body joints.
We follow the cross-view protocol \cite{shahroudy2016ntu}.
Following \cite{katsman2023riemannian}, we model each sequence as a $75 \times 75$ covariance matrix.

\subsection{Implementation Details}
As reported in \cref{sec:experiments}, MUL shows the best performance. Therefore, we view $A$ in \cref{eq:rw_mul_mlog} as the parameter for all experiments. In the following, we discuss in detail the specific implementation of each method.

\textbf{LieBN:}
We follow the official code\footnote{https://github.com/GitZH-Chen/LieBN} to implement the experiments.
The learning rate is $5e^{-2}$.
Since our LieBN-ALEM shows early convergence, we set the training epochs as 150, 50, and 30 for [93, 30], [93, 70, 30], and [93, 70, 50, 30] architectures.
Other settings are the same as \cref{sec:experiments}.

\textbf{RResNet:}
We follow the official code\footnote{https://github.com/CUAI/Riemannian-Residual-Neural-Networks} to implement the experiments.
For the HDM05 dataset, we use the Riemannian SGD \cite{becigneul2018riemannian} with a $5e^{-2}$ learning rate and a training epoch of 200.
For the NTU60 dataset, we use the Riemannian AMSGrad \cite{becigneul2018riemannian} with a $1e^{-2}$ learning rate and a training epoch of 50.
We adopt the architectures of [93, 30] and [75, 30] on these two datasets.

\textbf{Gyro MLR:}
Since the code of gyro MLR is not publicly available, we carefully re-implement the gyro MLR in \cite{nguyen2023building}. 
We adopt an architecture of [75, 30] under an SGD optimizer. The batch size and training epoch are 30 and 200, respectively.

\section{Proofs} \label{app:proofs}

\begin{proof} [Proof of \cref{thm:rethk_lem_lcm}]
Let us first deal with the $\biparamLEM$.
Putting the differential of matrix logarithm into \cref{eq:pullback_metrics}, one can directly obtain the result.

Now, let us focus on LCM.
Denote $\glcm$, $\geuc$, and $\gcm$ as LCM, standard Euclidean metric, and the metric on the Cholesky manifold \cite{lin2019riemannian}, respectively.
By \cref{eq:lcm}, $\{\spd{n}, \glcm\}$ is isometric to $\{\cho{n},\tilde{g}\}$, with Cholesky decomposition $\scrL$ as an isometry.
This is exactly how \cite{lin2019riemannian} derived LCM.
So, the key point lies in the Cholesky metric $\tilde{g}$.
Let us reveal why it is defined in this way.
In fact, $\tilde{g}$ is derived from $\geuc$ by $\clnchart$.
Simple computations show that
\begin{equation} \label{eq:diff_cho_ln_chart}
    \varphi_{ln *,L}(V) = \lfloor V \rfloor + \mathbb{D}^{-1}(L)\mathbb{D}(V),
\end{equation}
where $V \in T_L\cho{n}$.
By \cref{eq:diff_cho_ln_chart}, \cref{eq:lcm} can be rewritten as
\begin{equation}
    \gcm_L(X,Y) = \geuc(\varphi_{ln *,L}((X),\varphi_{ln *,L}((Y)).
\end{equation}
Therefore, $\clnchart: \cho{n} \rightarrow \tril{n}$ is an isometry.
By transitivity, $\cln: \spd{n} \rightarrow \tril{n}$ is also an isometry.
\end{proof}

\begin{proof}[Proof of \cref{cor:biparamLEM_pem}]
    As $\bbR{n(n+1)/2} \cong \tril{n} \cong \sym{n}$, LCM is therefore a pullback metric from the standard Euclidean space $\sym{n}$.
    Secondly, as every Euclidean space is naturally isometric, $\biparamLEM$ is therefore also a pullback metric from the standard Euclidean space $\sym{n}$.
\end{proof}

\begin{proof} [Proof of \cref{lem:g_spd}]
    By the definition of \cref{eq:phi_mul,eq:phi_sca_mul,eq:phi_innerpro,eq:phi_g}, \cref{enum:hilbert,enum:isomorphism} can be directly obtained.
    Now, let us focus on \cref{enum:mlog_riem_spd}.
    As every Euclidean space is an Abelian Lie group, $\{\spd{n}, \phiMul \}$ is an Abelian Lie group. 
    The geodesic distance in \cref{eq:dist_mlog} is also obvious, as $\phi$ is a Riemannian isometry.
    
    We only need to prove \cref{eq:gene_rie_exp_spd,eq:gene_rie_log_spd,eq:gene_pt_spd}.
    Note that in Euclidean space $\bbR{n}$, for any $x,y \in \bbR{n}$ and tangent vector $v \in T_x\bbR{n} \cong \bbR{n}$, we have the following
    \begin{align}
        \rieexp_x v &= x+v,\\
        \rielog_x y &= y-x,\\
        \pt{x}{y} v &= v.
    \end{align}
    By the isometry of $\phi$, we can readily obtain \cref{eq:gene_rie_exp_spd,eq:gene_rie_log_spd,eq:gene_pt_spd}.
\end{proof}
\begin{proof} [Proof of \cref{props:diffeo_mlog}]
    Obviously, $\mgexp$ is the inverse of $\mlog$. 
    What followed is to verify the smoothness of $\mlog$ and its inverse.
    
    According to Theorem 8.9 in \cite{magnus2019matrix}, the map producing an eigenvalue or an eigenvector from a real symmetric matrix is $\cinf$.
    Recalling $\mlog$ and its inverse map $\mgexp$, it's obvious that they are comprised of arithmetic calculation or composition of some smooth maps.
    Therefore, $\mlog$ ($\mgexp$) is a diffeomorphism.
\end{proof}

\begin{proof} [Proof of \cref{thm:mlog_spd_properties}]
    This is a direct result of \cref{lem:g_spd}.
\end{proof}

\begin{proof} [Proof of \cref{props:diff_mgexp_mlog}]
    The differentials of $\mgexp$ and $\mlog$ can be derived similarly. 
    In the following, we only present the process of deriving the differential of $\mlog$. 
    
    First, Let us recall the differentials of eigenvalues and eigenvectors.
    Theorem 8.9 in \cite{magnus2019matrix} offers their Euclidean differentials, which are the exact formulations for differentials under the canonical base on SPD manifolds.
    So, we can readily obtain the differentials of eigenvalues and eigenvectors as the following:
    \begin{align}
        \label{eq: diff_eig_value} \sigma_{*,S} (V) &= u^{\top} V u,\\
        \label{eq: diff_eig_vec} u_{*,S} (V) &= (\sigma I- S)^{+} V u,
    \end{align}
    where $S u = \sigma u, u^\top u = 1$, and $()^+$ is the Moore–Penrose inverse.

    By the RHS of \cref{eq:rw_org_mlog}, the differential map of $\mlog$ is
    \begin{equation} \label{eq:diff_mlog}
    \begin{aligned}
        &\diffmlog{S}(V) \\
        &= U_{*,S} (V) \log(\Sigma)U^\top + U (\log{\Sigma})_{*,S} (V) U^\top \\
        & + U \log(\Sigma) U^\top_{*,S} (V) \\
        &= Q+Q^\top + U^\top (\log{\Sigma})_{*,S} (V) U,
    \end{aligned}
    \end{equation}
    where $Q= U_{*,S} (V) \log(\Sigma)U^\top$.

    For the differential of diagonal logarithm, it is 
    \begin{equation} \label{eq: diff_ln}
        \log_{*,S}{\Sigma} = A \frac{1}{\Sigma} \Sigma_{*,S},
    \end{equation}
    where $A$ is defined in \cref{eq:rw_mul_mlog}.

    Denote the eigenvectors and eigenvalues of $S=U\Sigma U^\top$ as $U = (u_1,\cdots,u_n)$ and $\Sigma=\diag(\sigma_1,\cdots,\sigma_n)$.
    By \cref{eq: diff_eig_value}-\cref{eq: diff_ln}, the differential of $\mlog$ can be obtained.
\end{proof}
\begin{proof} [Proof of \cref{props:diff_mgexp_series}]
    Following the notations in the proposition, we make the following proof.
    By abuse of notation, in the following, we omit the wide tilde $\widetilde{~}$.
   
    Now, we proceed to deal with the differential of $\mgexp$.
    We rewrite the formula of $\mgexp$ as
    \small
    \begin{align}
        & \mgexp(X) \\
        &= U \balpha(\Sigma) U^\top,\\
        &= U \diag(e^{\ln^{a_1}\sigma_1},\cdots,e^{\ln^{a_n}\sigma_n})U^\top,\\
        &= U \diag(\sum_{k=0}^{\infty}\frac{(\ln^{a_1}\sigma_1)^k}{k!},\cdots,\sum_{k=0}^{\infty}\frac{(\ln^{a_n}\sigma_n)^k}{k!})U^\top,\\
        \label{eq:rw_mgexp_last2} &= U (\sum_{k=0}^{\infty}\frac{B\Sigma}{k!}) U^\top,\\
        \label{eq:rw_mgexp} &= \sum_{k=0}^{\infty}\frac{PX}{k!}
    \end{align}
    \normalsize
     where $P=UBU^\top$, with $U$ from eigendecomposition $X=U\Sigma U^\top$ and diagonal matrix $B=\diag(\ln^{a_1},\cdots,\ln^{a_n})$.
     By the properties of normed vector algebras \cite[Prop.~15.14]{loring2011introduction}, we can obtain the last equation.
     Then, we can compute the differential of $\mgexp$ by curves.
     Given a curve $c$ on $\sym{n}$ starting at $X$ with initial velocity $W \in T_X\sym{n}$, we have
     \begin{equation}
        \begin{aligned}
            \diffmgexp{X}(W) 
            &= \left. \frac{d}{dt} \right |_{t=0} \mgexp \circ c(t)\\
            \label{eq:diff_mgexp_series_proof_stp2}
            &= \left. \frac{d}{dt} \right |_{t=0}\sum_{k=0}^{\infty}\frac{Pc(t)}{k!}.
        \end{aligned} 
    \end{equation}
    By a term-by-term differentiation, we have
    \begin{equation} 
        \begin{aligned} \label{eq:diff_last2step_mgexp}
            &\diffmgexp{X}(W)\\ 
            &= \sum_{k=1}^{\infty} \frac{1}{k !}(\sum_{l=0}^{k-1}(PX)^{k-l-1} \left. \frac{d}{dt} \right|_{t=0}(Pc) (PX)^l).
        \end{aligned}
    \end{equation}
    By the chain rule, we have
    \begin{equation} \label{eq:chain_rule_Pc}
        \left. \frac{d}{dt} \right|_{t=0}(Pc) = P'(0)X + PV.
    \end{equation}
    $P'(0)$ is obtained by
    \begin{equation}
    \label{eq:diff_P_at0}
        \begin{aligned}
            P'(0) 
            &= (UBU^\top)'(0)\\
            &= U'(0)BU^\top + UBU^{\top '}(0)\\
            &= D_U B U^\top + U B D_U^\top,
        \end{aligned}
    \end{equation}

    where $D_U$ is derived from the differential of eigenvectors,
    \begin{equation} \label{eq:D_U}
    D_U = (\begin{array}{ccc}
             (\sigma_1 I-S)^+ V u_1 & \cdots & (\sigma_n I-S)^+ V u_n
        \end{array}).
    \end{equation}
    Applying \cref{eq:chain_rule_Pc}, \cref{eq:diff_P_at0} and \cref{eq:D_U} into \cref{eq:diff_last2step_mgexp}, we have the differential of $\mgexp$.
\end{proof}

\begin{proof} [Proof of \cref{props:geo_mean_spd}]
    Obviously, the metric space $\{\spd{n},\dalem\}$ is isometric to the space $\sym{n}$ endowed with the standard Euclidean distance.
    Therefore, the weighted Fréchet mean of $\{S_i\}$ in $\spd{n}$ corresponds to the weighted Fréchet mean of associated points $\{\mlog(S_i)\}$ in $\sym{n}$.
    The weighted Fréchet means in Euclidean spaces are clearly the familiar weighted means.
\end{proof}

\begin{proof} [Proof of \cref{props:biinvariance}]
    As $\mlog$ is a Riemannian isometry and $\sym{n}$ is bi-invariant, ALEM is therefore bi-invariant.
\end{proof}

\begin{proof} [Proof of \cref{props:exp_invariance}]
    Following the notations in this proposition, we make the following proof.
    The LHS can be rewritten as 
    \begin{equation}
        \begin{aligned}
            (\mathrm{FM}(S_1^\beta,\cdots S_m^\beta)) 
            &= \mgexp(\sum_{i=1}^{m} \frac{1}{m}\beta\mlog(S_i))\\
            &= \mgexp(\beta\sum_{i=1}^{m} \frac{1}{m}\mlog(S_i))\\
            &= [\mgexp(\sum_{i=1}^{m} \frac{1}{m}\mlog(S_i))]^\beta\\
            &= (\mathrm{FM}(S_1,\cdots S_m))^\beta.
        \end{aligned}
    \end{equation}
\end{proof}

\begin{proof}[Proof of \cref{props:frechet_means_add_props}]
    Recalling \cref{eq:fm_alem}, U\ref{props:u1} and U\ref{props:u2} obviously hold.
    
    When SPD matrices $\{A_i\}_{i \leq n}$ commutes, we have
    \begin{equation} \label{eq:fm_commute_alem}
        \fm(\{A_i\})= (\sum A_i)^{\frac{1}{n}}.
    \end{equation}
    With \cref{eq:fm_commute_alem}, V\ref{props:v1}-V\ref{props:v4} can be easily proved.
\end{proof}

\begin{proof} [Proof of \cref{props:sim_invariance}]
    Obviously, for a given SPD matrix $S$,
    \begin{align}
        \label{eq:mlog_rotate} \mlog(R S R^\top) &= R \mlog(S) R^\top,\\
        \label{eq:mlog_scale} \mlog(s^2 S ) &=  U(\log(s^2 I)+\mlog(\Sigma))U^\top,
    \end{align}
    where $S=U \Sigma U\top$ is the eigendecomposition.
    We can obtain the results with \cref{eq:mlog_rotate} and \cref{eq:mlog_scale}.
\end{proof}

\begin{proof}[Proof of \cref{prop:rewrit_general_log}]
    The three equations can be directly obtained.
\end{proof}

\begin{proof} [Proof of \cref{props:grad_mlog}]
    \cref{eq:gradient_eigen_function} is the so-called Daleck\u ii-Kre\u in formula presented in \cite[P. 60]{bhatia2009positive}.
    Now, let us focus on the gradient w.r.t $A$.
    Differentiating both sides of \cref{eq:rw_mul_mlog}:
    \begin{equation}
        \diff X = (*) + U \diff A \odot \log(\Sigma)U^T,
    \end{equation}
    where $(*)$ means other parts related to $\diff U$ and $\diff \Sigma$.
    According to the invariance of first-order differential form, we have,
    \begin{align}
        & \nabla_{X} L : \diff X \nonumber \\ 
        &= \nabla_{S} L:\diff S + \nabla_{X} L: (U \diff A \odot \log(\Sigma)U^T) \\
        \label{eq:diif_last_row}
        &= \nabla_{S}L :\diff S + [U^\top (\nabla_{X} L) U] \odot \log(\Sigma):\diff A,
    \end{align}
    where $A:B=\tr(A^\top B)$ is the Euclidean Frobenius inner product.
    From the second term on the RHS of \cref{eq:diif_last_row}, we can obtain the gradient w.r.t $A$.
\end{proof}

\begin{proof}[Proof of \cref{props:grad_mexp}]
    The derivation follows the same logic as \cref{props:grad_mlog}.
    We only need to show the derivation of \cref{eq:gradient_mexp_wrt_A}.
    Similar with \cref{props:grad_mlog}, we have the following:
    \begin{equation}
        \diff X = (*) + U \diff A \odot \left( \balpha (X) \frac{-\Sigma}{A^2} \right) U^T,
    \end{equation}
    \begin{align}
    &\nabla_{X} L : \diff X \nonumber \\
    &= \nabla_{S}L :\diff S + [U^\top (\nabla_{X} L) U] \odot \left( \balpha (X) \frac{-\Sigma}{A^2} \right):\diff A.
    \end{align}    
\end{proof}

\begin{proof}[Proof of \cref{thm:gyro_mlr_alem}]
    
    Following \cite{nguyen2022gyro,nguyen2023building}, we first define gyro structures under ALEM:
    \begin{align}
        \label{eq:gyro_addtion}
        P \oplus Q &= \rieexp_{P}\left(\pt{E}{P} \left(\rielog _{E}(Q)\right)\right), \\
        \label{eq:gyro_automorphism}
        \gyr[P, Q] R &= (\ominus(P \oplus Q)) \oplus(P \oplus(Q \oplus R)),\\
        \label{eq:gyro_scalar_product}
        t \otimes P &= \rieexp_{E}\left(t \rielog _{E}(P)\right), \\
        \label{eq:gyro_inverse}
        \ominus P &= -1 \otimes P = \rieexp_{E}\left(- \rielog _{E}(P)\right), \\
        \label{eq:gyro_inner_product}
        \gyrinner{P}{Q}&=\left\langle \rielog _I (P), \rielog _I (Q)\right\rangle_{I},\\
        \label{eq:gyro_norm}
        \gyrnorm{P} &= \gyrinner{P}{P},\\
        \label{eq:gyro_distance}
        \gyrdist(P, Q) &= \gyrnorm{\ominus P \oplus Q},
    \end{align}
    where $P,Q,R \in \spd{n}$, and $I$ is the identity matrix.
    The above operations are called gyro addition, gyro automorphism, gyro scalar product, gyro inverse, gyro inner product, gyro norm, and gyrodistance. Simple computations show that \cref{eq:gyro_addtion} and \cref{eq:gyro_scalar_product} are the exact $\mlogMul$ and $\mlogMulScalar$ in \cref{thm:mlog_spd_properties}.
    As indicated by \cref{thm:mlog_spd_properties}, $\{\spd{n}, \mlogMul, \mlogMulScalar \}$ forms a gyro vector space \cite[Def. 1]{nguyen2022gyrovector}.
    In the following proof, we follow the notations in \cite{nguyen2023building} to use $\odot$ and $\oplus$.

    The gyro MLR \cite{nguyen2023building} under ALEM is defined as 
    \begin{equation}
        \label{eq:mlr_alem}
        \begin{aligned}
        &p(y=k \mid S \in \spd{n}) \\
        &\propto \exp \left(\operatorname{sign}(\langle \tilde{A}_k, \rielog_{P_k}(S) \rangle_{P_k})\|\tilde{A}_k\|_{P_k} \bar{d} (S, H_{\tilde{A}_k, P_k}) \right),\\
    \end{aligned}
    \end{equation}
    where $P_k \in \spd{n}$ and $\tilde{A}_k \in T_{P_k} \spd{n}$.
    $\bar{d} (S, H_{\tilde{A}_k, P_k})$ is the margin distance to the SPD hyperplane $H_{\tilde{A}_k, P_k}$, which is defined as
    \begin{align}
        \label{eq:gyro_dist_hyperplane}
        \bar{d} (S, H_{\tilde{A}_k, P_k}) &=\sin (\angle S P_k Q^*) \gyrdist(S,P_k), \\
        Q^* &=\underset{Q \in H_{P_k, \tilde{A}_k} \backslash \{P_k\}}{\arg \max }\left(\cos (\angle S P_k Q)\right), \\
        \label{eq:gyro_consie}
        \cos (\angle S P_k Q) &= \frac{\gyrinner{ \ominus P_k \oplus Q}{\ominus P_k \oplus S}}{\gyrnorm{ \ominus P_k \oplus Q} \gyrnorm{ \ominus P_k \oplus S }},\\
        \label{eq:gyro_hyperplane}
        H_{\tilde{A}_k, P_k} &= \{S \in \spd{n}: \langle \rielog_{P_k} S, \tilde{A}_k \rangle_{P_k} =0\}.
    \end{align}
    \cref{eq:gyro_dist_hyperplane,eq:gyro_consie,eq:gyro_hyperplane} are called the SPD Pseudo-gyrodistance, SPD gyrocosine, and SPD hypergyroplane.

    For simplicity, we further omit the subscript $k$ in $P_k$ and $\tilde{A_k}$.
    \cref{eq:gyro_hyperplane} can be simplified:
    \begin{equation}
        \label{eq:alem_gyro_hyperplane_simplified}
        \begin{aligned}
            &\langle \rielog_{P} S, \tilde{A} \rangle_{P}\\
            &\stackrel{(1)}{=} \left\langle (\diffmlog{P})^{-1} (\mlog(P)-\mlog(S)) , \tilde{A} \right \rangle_{P} \\
            &\stackrel{(2)}{=} \left\langle \diffmlog{P} \circ (\diffmlog{P})^{-1} (\mlog(P)-\mlog(S)) , \diffmlog{P} \tilde{A} \right\rangle \\
            &= \left\langle \mlog(P)-\mlog(S) , \diffmlog{P} (\tilde{A}) \right\rangle.
        \end{aligned}
    \end{equation}
    The above derivation comes from the following.
    \begin{enumerate}[(1)]
        \item
        \cref{eq:rielog_gmlog}.
        \item 
        The definition of ALEM.
    \end{enumerate}

    Similarly, simple computation shows that \cref{eq:gyro_consie} can also be simplified as
    \begin{equation} \label{eq:alem_gyro_consine_simplified}
        \frac{\left \langle -\mlog(P)+\mlog(Q), -\mlog(P)+\mlog(S) \right \rangle}{\|  -\mlog(P)+\mlog(Q)\|_\rmF \|-\mlog(P)+\mlog(S) \|_\rmF}.
    \end{equation}

    Combined with \cref{eq:alem_gyro_hyperplane_simplified,eq:alem_gyro_consine_simplified}, \cref{eq:gyro_dist_hyperplane} is equivalent to the distance to the hyperplane in the Euclidean space.
    Therefore, \cref{eq:gyro_dist_hyperplane} has a closed form solution:
    \begin{equation}
    \label{eq:dist_hyperplane_final}
    \begin{aligned}
        &\bar{d} (S, H_{\tilde{A}, P})\\
        &= \frac{\left|\left \langle \mlog(S)- \mlog(P), \bar{A} \right \rangle \right|}{\left \| \bar{A} \right \|_\rmF}\\
        &= \frac{\left|\left \langle \mlog(S)- \mlog(P), \bar{A} \right \rangle \right|}{\left \|  A \right \|_P},
    \end{aligned}
    \end{equation}
    where $\bar{A}=\diffmlog{P}(\tilde{A})$.
    Putting \cref{eq:dist_hyperplane_final} into \cref{eq:mlr_alem}, one can get the results.
\end{proof}

\begin{proof} [Proof of \cref{propos:param_by_geom}]
    Let's first review the update formulation in the RSGD \cite{bonnabel2013stochastic}, which is, geometrically speaking, a natural generalization of Euclidean stochastic gradient descent.
    For a minimization parameter $w$ on an $n$-dimensional smooth connected Riemannian manifold $\calM$, we have the following update, 
    \begin{equation} \label{eq:riem_sgd}
        w^{(t+1)}=\rieExp{w^{(t)}} (-\gamma^{(t)} \pi_{w^{(t)}}(\nabla_{w^{(t)}} L )),
    \end{equation} 
    where $\rieExp{w}(\cdot): T_w\calM \rightarrow \calM$ is the Riemannian exponential map, which maps a tangent vector at $w$ back into the manifold $\calM$, and $\pi_{w}(\cdot): \bbR{n} \rightarrow T_w\calM$ is the projection operator, projecting an ambient Euclidean vector into the tangent space at $w$.
    In the case of the SPD manifold, $\forall S \in \spd{n}, \forall X \in \bbR{n \times n}, \forall V \in \sym{n}$, the exponential map and projection operator is formulated as the following:
    \begin{align}
        \label{eq:spd_proj} \pi_{S}(X) &= S \frac{X+X^\top}{2}S,\\
        \label{eq:spd_exp} \rieExp{S}(V) &= S^{1/2} \mexp(S^{-1/2} V S^{-1/2}) S^{1/2},
    \end{align}
    where $\mexp(\cdot)$ is the matrix exponential.
    For more details about \cref{eq:spd_proj} and \cref{eq:spd_exp}, please kindly refer to \cite{yger2013review} and \cite{amari2016information}.
    Substitute \cref{eq:spd_proj} and \cref{eq:spd_exp} into \cref{eq:riem_sgd}, \cref{eq:update_pos_scalar} can be immediately obtained.
\end{proof}
\begin{proof} [Proof of \cref{props: geom_equivalent_div}]
    Without loss of generality, we focus on the equivalence between $b=B_{11}$ and $a=\alpha_{11}$.
    Let us denote $\log_{e}^{(\cdot)}$ as $\ln^{(\cdot)}$.
    Note that $b$ is essentially expressed as $b=\ln^a$.
    Supposing $b^{(t)} = \ln^{a^{(t)}}$, then we have
    \begin{equation}
        \begin{aligned}
            \nabla_{a^{(t)}} L 
            &= \nabla_{b^{(t)}} L \frac{\partial \ln^a }{a} |_{a^{(t)}}\\
            &= \nabla_{b^{(t)}} L \frac{1}{a^{(t)}}.
        \end{aligned}
    \end{equation}

    By \cref{eq:update_pos_scalar}, $\ln^{a^{(t+1)}}$ is 
    \begin{equation}
        \begin{aligned}
            \ln^{a^{(t+1)}} 
            &= \ln^{a^{(t)}e^{-\gamma^{(t)} a^{(t)}  \nabla_{a^{(t)}} L}}\\
            &= \ln^{a^{(t)}} -\gamma^{(t)} a^{(t)}  \nabla_{a^{(t)}} L\\
            &= \ln^{a^{(t)}} - \gamma^{(t)} a^{(t)} (\nabla_{b^{(t)}} L/a^{t}) \\
            &= \ln^{a^{(t)}} -\gamma^{(t)}\nabla_{b^{(t)}} L\\
            &= b^{t} -\gamma^{(t)}\nabla_{b^{(t)}} L.
        \end{aligned}
    \end{equation}
    The last row is the updated formula of ESGD for $b$.
    
    Therefore, supposing $b^{(0)} = \ln^{a^{(0)}}$, then the optimization results after the overall training are equivalent.
\end{proof}

\end{document}